\documentclass[11pt]{article}

\usepackage{acl}

\usepackage{times}
\usepackage{latexsym}
\usepackage[T1]{fontenc}
\usepackage[utf8]{inputenc}
\usepackage{microtype}
\IfFileExists{inconsolata.sty}{\usepackage{inconsolata}}{}
\usepackage{graphicx}

\usepackage{xcolor}
\usepackage{tikz}
\usetikzlibrary{shadows.blur}
\usetikzlibrary{calc}
\usepackage[edges]{forest}
\usepackage{adjustbox}
\usepackage{enumitem}
\usepackage{booktabs}
\usepackage{tabularx}
\usepackage{amssymb}
\usepackage{multirow}
\usepackage{cuted}
\newcolumntype{L}{>{\raggedright\arraybackslash}X}

\newcommand{\hbfull}{\ensuremath{\bullet}}
\newcommand{\hbhalf}{\ensuremath{\circledcirc}}
\newcommand{\hbnone}{\ensuremath{\circ}}

\newcommand{\bcref}[1]{\hyperref[sec:boundary-protocol]{(#1)}}

\definecolor{rootfg}{RGB}{55,55,75}

\definecolor{edgeblue}{RGB}{85,140,195}
\definecolor{fillblue1}{RGB}{50,110,170}
\definecolor{fillblue2}{RGB}{130,172,210}
\definecolor{fillblue3}{RGB}{178,206,232}
\definecolor{fillblue4}{RGB}{220,234,248}

\definecolor{edgegreen}{RGB}{65,145,65}
\definecolor{fillgreen1}{RGB}{45,125,50}
\definecolor{fillgreen2}{RGB}{120,182,125}
\definecolor{fillgreen3}{RGB}{172,212,175}
\definecolor{fillgreen4}{RGB}{216,238,218}

\definecolor{edgeyellow}{RGB}{170,140,25}
\definecolor{fillyellow1}{RGB}{160,128,10}
\definecolor{fillyellow2}{RGB}{200,183,100}
\definecolor{fillyellow3}{RGB}{228,218,160}
\definecolor{fillyellow4}{RGB}{248,243,215}

\definecolor{edgeorange}{RGB}{185,110,45}
\definecolor{fillorange1}{RGB}{178,95,30}
\definecolor{fillorange2}{RGB}{215,162,110}
\definecolor{fillorange3}{RGB}{235,202,168}
\definecolor{fillorange4}{RGB}{250,234,218}

\definecolor{edgeviolet}{RGB}{130,80,165}
\definecolor{fillviolet1}{RGB}{110,60,145}
\definecolor{fillviolet2}{RGB}{165,130,195}
\definecolor{fillviolet3}{RGB}{200,178,220}
\definecolor{fillviolet4}{RGB}{232,222,242}

\tikzset{
  rootnode/.style={
    draw=rootfg, line width=1.0pt, rounded corners=4pt,
    align=center, font=\footnotesize\bfseries,
    inner xsep=6pt, inner ysep=5pt,
    fill=white, text=rootfg,
    blur shadow={shadow blur steps=6,shadow xshift=0.5pt,shadow yshift=-0.5pt,shadow blur radius=1.4pt}
  },
  tracknode/.style={
    draw=none, rounded corners=3pt,
    align=center, font=\fontsize{8.5}{9.2}\selectfont\bfseries,
    minimum height=1.7em, text width=8.0em,
    inner xsep=2.2pt, inner ysep=1.4pt,
    text=white,
    blur shadow={shadow blur steps=5,shadow xshift=0.4pt,shadow yshift=-0.4pt,shadow blur radius=1.0pt}
  },
  familynode/.style={
    draw=none, rounded corners=2.5pt,
    align=center, font=\fontsize{8.0}{8.6}\selectfont\bfseries,
    minimum height=1.5em, text width=8.4em,
    inner xsep=1.8pt, inner ysep=1.0pt
  },
  micronode/.style={
    draw=none, rounded corners=2pt,
    align=center, font=\fontsize{8.0}{8.6}\selectfont,
    minimum height=1.5em, text width=9.8em,
    inner xsep=1.6pt, inner ysep=0.9pt
  },
  leafnode/.style={
    draw=none, rounded corners=1.5pt,
    align=left, font=\fontsize{8.0}{8.6}\selectfont,
    text width=17.5em,
    inner xsep=2.0pt, inner ysep=1.0pt
  }
}

\newcommand{\ml}[2]{#1~\cite{#2}}
\newsavebox{\taxonomytreebox}
\newsavebox{\taxonomytreeboxFULL}
\newsavebox{\whenadapttreebox}

\title{Unsupervised Post-Training of Foundation Models: A Survey}

\author{
  \textbf{Yijie Xu$^{1}$ \quad Qianyi Cai$^{1}$ \quad Huizai Yao$^{1}$ \quad
  Yili Wang$^{1}$ \quad Tianfu Wang$^{1}$ \quad Cehao Yang$^{1}$} \\[0.35em]
  \textbf{Xingbo Yao$^{1,3}$ \quad Zhiyu Guo$^{3}$ \quad Aiwei Liu$^{4}$ \quad
    Xuming Hu$^{1,2,\dagger}$ \quad Weiyu Guo$^{5,6,\dagger}$ \quad
  Hui Xiong$^{1,2,\dagger}$} \\[0.7em]
  \small
  $^{1}$HKUST(GZ) \quad $^{2}$HKUST \quad $^{3}$Xiaohongshu Inc. \quad
  $^{4}$WeChat, Tencent \quad $^{5}$CUHK \quad $^{6}$AI$^{2}$ Robotics \\
  $^{\dagger}$\,Corresponding authors.
}

\begin{document}
\maketitle
\vspace{-3.2\baselineskip}

\begin{lrbox}{\taxonomytreebox}
  \begin{tikzpicture}[
      every node/.style={align=center, rounded corners=3pt,
      inner xsep=6pt, inner ysep=5pt, font=\footnotesize},
      qbox/.style={draw=rootfg, line width=0.8pt, fill=white,
      text=rootfg, font=\footnotesize\itshape, text width=0.78\linewidth},
      fbox/.style={draw=none, text=white, font=\footnotesize\bfseries,
        text width=0.44\linewidth, minimum height=4.0em,
      inner xsep=3pt},
      abox/.style={draw=edgeviolet, line width=0.8pt, dashed,
        fill=fillviolet4, text=fillviolet1, font=\footnotesize\bfseries,
        text width=0.90\linewidth, minimum height=2.6em,
      inner xsep=3pt},
    ]
    \node[qbox] (Q) at (0,0)
    {What internal update object is used\\to compute the update signal?};
    \node[fbox, fill=fillblue1, anchor=north east]
    at ([yshift=-1.2ex,xshift=-0.15em]Q.south) (F1)
    {Prediction-Statistic\\Optimization\\
      {\fontsize{6.8}{7.6}\selectfont\mdseries\itshape NLL, entropy, confidence}\\
    {\fontsize{6.8}{7.6}\selectfont\mdseries e.g.,\ \textsc{CPT-LM}, \textsc{EM-RL}}};
    \node[fbox, fill=fillgreen1, anchor=north west]
    at ([yshift=-1.2ex,xshift=0.15em]Q.south) (F2)
    {Sample-Relation\\Supervision\\
      {\fontsize{6.8}{7.6}\selectfont\mdseries\itshape majority vote, semantic clusters}\\
    {\fontsize{6.8}{7.6}\selectfont\mdseries e.g.,\ \textsc{TTRL}, \textsc{EMPO}}};
    \node[fbox, fill=fillyellow1, anchor=north]
    at ([yshift=-0.7ex]F1.south) (F3)
    {Self-Generated Target\\Bootstrapping\\
      {\fontsize{6.8}{7.6}\selectfont\mdseries\itshape pseudo-labels, rationales, curricula}\\
    {\fontsize{6.8}{7.6}\selectfont\mdseries e.g.,\ \textsc{Self-Improve}, \textsc{R-Zero}}};
    \node[fbox, fill=fillorange1, anchor=north]
    at ([yshift=-0.7ex]F2.south) (F4)
    {Internal Evaluator\\Bootstrapping\\
      {\fontsize{6.8}{7.6}\selectfont\mdseries\itshape self-judge, reward model, rubric}\\
    {\fontsize{6.3}{7.0}\selectfont\mdseries e.g.,\ \mbox{\textsc{Self-Rewarding LM}}, \textsc{CoNL}}};
    \node[abox, anchor=north]
    at ([yshift=-0.7ex]$(F3.south)!0.5!(F4.south)$)
    {Adjacent Methods (boundary)\\
      {\fontsize{6.8}{7.6}\selectfont\mdseries\itshape no-update inference-time; verifier/tool-assisted; human/seed-supervised; stronger-teacher; external-evaluator}\\
    {\fontsize{6.8}{7.6}\selectfont\mdseries e.g.,\ \textsc{EM-INF}, \textsc{T$^3$RL}, \textsc{Absolute Zero}}};
  \end{tikzpicture}
\end{lrbox}

\begin{abstract}
  \vspace{-0.25em}
  Foundation-model post-training usually relies on human labels,
  preference data, stronger teachers, or executable verifiers. We
  study \emph{Unsupervised Post-Training} (UPT): update-bearing
  adaptation on unlabeled inputs whose learning signal is derived
  from same-lineage model artifacts rather than an external oracle.
  We catalog 80 strict UPT methods and organize them by the object
  that supplies the update signal: a prediction statistic, a sample
  relation, a self-generated target, or an internal evaluator. Beyond
  inventory, we show how the choice of internal signal and task
  structure determines whether post-training improves the model or
  recursively amplifies error. An orthogonal
  Input~Visibility~$\times$~Update~Persistence view maps deployment
  regimes and defines a unified framework for UPT selection and
  evaluation.
\end{abstract}

\section{Introduction}

Foundation-model post-training has so far followed two waves of
external supervision.\footnote{We use \emph{foundation models}
  for both text-only and multimodal large models; a recurring
  finding is that the dominant unsupervised
mechanisms are modality-agnostic.}
The first wave is \emph{human labels}, including supervised
fine-tuning and preference-based RL. The second is \emph{external
verifiers}: math checkers, unit tests, and executable environments
that license RL with verifiable
rewards~\citep{shao2024deepseekmath,zhao2025absolutezero,liao2026t3rl}. A third
wave, accumulating since 2023 and accelerating through 2025--2026,
abandons external supervision and updates the model using
\emph{only} unlabeled prompts, text, or target inputs, drawing
every update signal from the model's own samples, distributions,
judges, or
curricula~\citep{huang2023selfimprove,yuan2024selfrewarding,zuo2025ttrl,huang2026rzero}.

\begin{figure}[t]
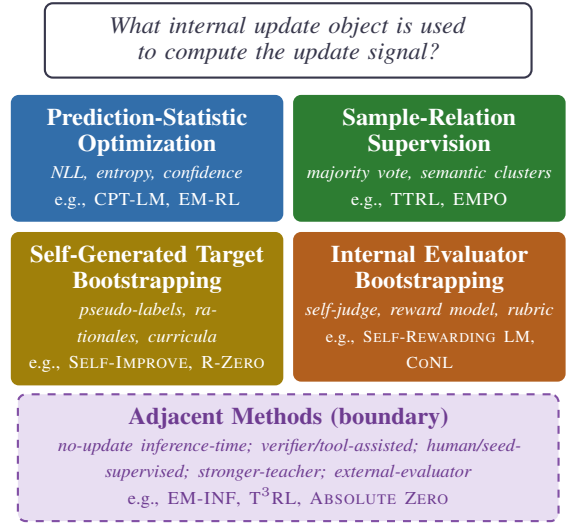

  \centering
  \begin{adjustbox}{max totalsize={0.98\linewidth}{0.32\textheight},center}
    \usebox{\taxonomytreebox}
  \end{adjustbox}
  \caption{Update-object taxonomy of \emph{Unsupervised
    Post-Training}. Strict UPT methods are grouped by the internal
    update object. Dashed boxes denote boundary-adjacent methods that
  fail at least one strict UPT check.}
  \vspace{-1em}
  \label{fig:taxonomy}
\end{figure}

\noindent
UPT enables adaptation when labels and task-specific verifiers
cannot be obtained or transferred. This setting covers newly
arrived domain corpora and open-ended generation tasks such as
dialogue and summarization, where exact-answer checkers are
unavailable.

We use \emph{UPT} for update-bearing adaptation whose learning
signal comes from same-lineage model artifacts;
\S\ref{sec:definitions} operationalizes this scope with four
boundary checks. Existing surveys cover neighboring territory:
LLM self-improvement~\citep{tao2024selfevolution,kumar2025posttraining},
test-time adaptation~\citep{liang2025ttasurvey}, and reinforced
reasoning~\citep{xu2025reasoningmodels,chen2025longcotsurvey}. None
of them, however, enforces the three constraints that
define our scope: a real update, no external supervision, and a
classification axis defined by the internal object that produces
the update signal. Appendix~\ref{app:survey-comparison} provides the
dimension-wise comparison.

We organize UPT by the internal object consumed by the update:
a prediction statistic, a sample relation, a self-generated target,
or an internal evaluator. Figure~\ref{fig:taxonomy} shows the four
resulting families: Prediction-Statistic Optimization,
Sample-Relation Supervision, Self-Generated Target Bootstrapping,
and Internal Evaluator Bootstrapping. A parallel adjacent track
maps methods that introduce verifier, tool, seed, teacher, or
external-evaluator signals. Together, the four families expose a
common error chain: an imperfect proxy selects or rewards outputs,
the update concentrates the model on them, and the next round reads
an even more biased proxy. The family determines where this loop
begins and which safeguard can interrupt it.

Our contributions are: (i) an operational boundary protocol that
distinguishes signal provenance from task structure
(\S\ref{sec:definitions}, Appendix~\ref{app:protocol});
(ii) an update-object taxonomy of four families and an orthogonal
Input Visibility~$\times$~Update Persistence view
(\S\ref{sec:family-i}--\S\ref{sec:timing}), with the
hierarchy and inventory cross-sections in
Appendix~\ref{app:inventory}; and (iii) a
cross-family synthesis of applicability, error propagation, and
deployment checks, supported by empirical and task-structure audits
(\S\ref{sec:synthesis}--\S\ref{sec:open-problems},
Appendix~\ref{app:synthesis-evidence}).\footnote{Companion
\href{https://github.com/yeahjack/awesome-unsupervised-post-training}{inventory}.}

\begin{lrbox}{\taxonomytreeboxFULL}
  \begin{forest}
    for tree={
      forked edges,
      grow=east,
      reversed=true,
      anchor=center,
      parent anchor=east,
      child anchor=west,
      base=center,
      font=\small,
      rounded corners,
      s sep=1.2pt,
      l sep=4.6pt,
      fork sep=3.2pt,
      inner xsep=1.4pt,
      inner ysep=0.7pt,
      edge+={rounded corners=1.4pt},
    },
    [
      Unsupervised Post-Training\\of Foundation Models,
      rootnode,
      rotate=90,
      anchor=north,
      [Prediction-Statistic\\Optimization, tracknode, fill=fillblue1,
        edge={draw=edgeblue, line width=0.9pt, rounded corners=2pt},
        for descendants={edge={draw=edgeblue, line width=0.7pt, rounded corners=1.5pt}},
        [Direct optimization, familynode, fill=fillblue2,
          [Predictive likelihood\\minimization, micronode, fill=fillblue3,
            [{\ml{CPT-LM}{ke2023cptlm}; \ml{Simple CPT}{qian2024simplescalable}; \ml{LangAdapt CPT}{elhady2025languageadapt}; \ml{Stability-Gap CPT}{guo2025stabilitygap}; \ml{ReplayAlign CPT}{abbes2025replayalign}; \ml{E2-LLM}{liu2024e2llm}; \ml{Data Eng 128K}{fu2024dataengineering}; \ml{LongContext Scaling}{xiong2024longcontext}; \ml{TLM}{hu2025test}; \ml{TTT-NN}{jang2024tttnn}; \ml{Long TTT}{bansal2026qttt}; \ml{In-Place TTT}{feng2026inplace}}, leafnode, fill=fillblue4]
          ]
          [Entropy / confidence\\minimization, micronode, fill=fillblue3,
            [{\ml{EM-FT}{agarwal2025unreasonable}; \ml{One-shot EM}{gao2025oneshot}}, leafnode, fill=fillblue4]
          ]
          [Sample-local state\\update, micronode, fill=fillblue3,
            [{\ml{SLOT}{hu2025slot}; \ml{SyTTA}{xu2025you}; \ml{Model Whisper}{kang2025modelwhisper}; \ml{ULDTTA}{xu2026uldtta}}, leafnode, fill=fillblue4]
          ]
        ]
        [Policy optimization, familynode, fill=fillblue2,
          [Entropy / confidence\\reward, micronode, fill=fillblue3,
            [{\ml{EM-RL(seq)}{agarwal2025unreasonable}; \ml{EM-RL(tok)}{agarwal2025unreasonable}; \ml{RENT}{prabhudesai2025rent}; \ml{RLSC}{li2025rlsc}}, leafnode, fill=fillblue4]
          ]
          [Geometric / rule-based\\statistics, micronode, fill=fillblue3,
            [{\ml{VIGOR}{wen2026verifierfreerlllmsintrinsic}; \ml{Latent-GRPO}{zhang2026silencejudgereinforcementlearning}; \ml{SSL-R1}{xie2026sslr1selfsupervisedvisualreinforcement}; \ml{SUDER$^{\ddagger}$}{hong2025suderselfimprovingunifiedlarge}}, leafnode, fill=fillblue4]
          ]
        ]
      ]
      [Sample-Relation\\Supervision, tracknode, fill=fillgreen1,
        edge={draw=edgegreen, line width=0.9pt, rounded corners=2pt},
        for descendants={edge={draw=edgegreen, line width=0.7pt, rounded corners=1.5pt}},
        [Self-consistency in\\one prompt, familynode, fill=fillgreen2,
          [Semantic / cluster\\consensus, micronode, fill=fillgreen3,
            [{\ml{EMPO}{zhang2025rightquestion}; \ml{Intuitor}{zhao2025intuitor}; \ml{CoVo}{zhang2025covo}; \ml{Co-rewarding}{zhang2026corewarding}}, leafnode, fill=fillgreen4]
          ]
        ]
        [Multi-sample consensus\\and test-time RL, familynode, fill=fillgreen2,
          [Majority-vote\\reward, micronode, fill=fillgreen3,
            [{\ml{TTRL}{zuo2025ttrl}; \ml{ETTRL}{liu2025ettrl}; \ml{ECHO}{zhao2026echo}; \ml{SPINE}{wu2025spine}; \ml{SCOPE}{wang2025scope}; \ml{COMPASS}{xing2025compass}; \ml{SCRL}{yan2026scrl}; \ml{RoiRL}{arzhantsev2025roirl}; \ml{EVOL-RL}{zhou2025evolrl}}, leafnode, fill=fillgreen4]
          ]
          [Wrong-majority\\corrections, micronode, fill=fillgreen3,
            [{\ml{Self-Harmony}{wang2026selfharmony}; \ml{DARE}{du2026dare}; \ml{RLCCF}{yuan2025rlccf}; \ml{Dual Consensus}{du2026dualconsensusescapingspurious}; \ml{CSRS}{yu2026stabilizingunsupervisedselfevolutionmllms}}, leafnode, fill=fillgreen4]
          ]
          [Multimodal\\consensus, micronode, fill=fillgreen3,
            [{\ml{EvoLMM}{thawakar2025evolmm}; \ml{TTRV}{singh2025ttrv}; \ml{MM-UPT}{wei2025mmupt}; \ml{EvoQuality}{wen2026selfevolvingvisionlanguagemodelsimage}}, leafnode, fill=fillgreen4]
          ]
        ]
      ]
      [Self-Generated Target\\Bootstrapping, tracknode, fill=fillyellow1,
        edge={draw=edgeyellow, line width=0.9pt, rounded corners=2pt},
        for descendants={edge={draw=edgeyellow, line width=0.7pt, rounded corners=1.5pt}},
        [Direct optimization, familynode, fill=fillyellow2,
          [Knowledge / instruction\\self-curation, micronode, fill=fillyellow3,
            [{\ml{Self-Tuning}{zhang2025selftuning}; \ml{KBAlign}{zeng2025kbalign}; \ml{CYCLE-INSTRUCT}{shen2025cycleinstruct}; \ml{LongMagpie}{gao2025longmagpie}}, leafnode, fill=fillyellow4]
          ]
          [Rationale / latent-thought\\self-training, micronode, fill=fillyellow3,
            [{\ml{Self-Improve}{huang2023selfimprove}; \ml{Quiet-STaR}{zelikman2024quietstar}; \ml{Confident ST}{wang2025confidentreasoning}; \ml{GENIUS}{xu2025genius}; \ml{LRM Self-Train}{shi2025lrmselftrain}; \ml{DTE}{liu2025dte}; \ml{Long Self-Improve}{wang2024longselfimprove}}, leafnode, fill=fillyellow4]
          ]
          [Curriculum / plan\\synthesis, micronode, fill=fillyellow3,
            [{\ml{TTCS}{yang2026ttcs}; \ml{DiSCTT}{moradi2026disctt}; \ml{TTSR}{he2026ttsr}; \ml{R-Zero}{huang2026rzero}; \ml{QueST}{song2026queryconditionedtesttimeselftraininglarge}; \ml{V-Zero}{wang2026vzeroselfimprovingmultimodalreasoning}}, leafnode, fill=fillyellow4]
          ]
        ]
        [Preference optimization, familynode, fill=fillyellow2,
          [Internally generated\\preference pairs, micronode, fill=fillyellow3,
            [{\ml{ScPO}{prasad2025scpo}; \ml{MACA}{samanta2025maca}; \ml{LongPO}{chen2025longpo}; \ml{RLSF}{vanniekerk2025rlsf}; \ml{G-Zero}{huang2026gzeroselfplayopenendedgeneration}}, leafnode, fill=fillyellow4]
          ]
        ]
      ]
      [Internal Evaluator\\Bootstrapping, tracknode, fill=fillorange1,
        edge={draw=edgeorange, line width=0.9pt, rounded corners=2pt},
        for descendants={edge={draw=edgeorange, line width=0.7pt, rounded corners=1.5pt}},
        [Preference optimization, familynode, fill=fillorange2,
          [Self-rewarding /\\meta-judge DPO, micronode, fill=fillorange3,
            [{\ml{Self-Rewarding LM}{yuan2024selfrewarding}; \ml{CREAM}{wang2024cream}; \ml{Meta-Rewarding}{wu2024metarewarding}; \ml{Temporal SRLM}{jin2025temporalselfrewarding}}, leafnode, fill=fillorange4]
          ]
        ]
        [Policy optimization, familynode, fill=fillorange2,
          [Evaluator-driven\\PG, micronode, fill=fillorange3,
            [{\ml{CoNL}{sui2026conl}; \ml{RLME}{rentschler2026rlme}; \ml{Meta-TTRL}{tan2026metattrl}; \ml{AERO}{gao2026aeroautonomousevolutionaryreasoning}; \ml{Self-Judge}{wu2026modelsjudgethemselvesunsupervised}; \ml{GvU$^{\ddagger}$}{pan2026learninggenerateunderstandingunderstandingdriven}}, leafnode, fill=fillorange4]
          ]
        ]
      ]
      [Adjacent Methods\\(boundary), tracknode, fill=fillviolet1,
        edge={draw=edgeviolet, line width=0.9pt, rounded corners=2pt, dashed},
        for descendants={edge={draw=edgeviolet, line width=0.7pt, rounded corners=1.5pt}},
        [No-update\\inference-time, familynode, fill=fillviolet2,
          [Logit / hidden-state\\descent, micronode, fill=fillviolet3,
            [{\ml{EM-INF}{agarwal2025unreasonable}}, leafnode, fill=fillviolet4]
          ]
        ]
        [Verifier- /\\tool-assisted, familynode, fill=fillviolet2,
          [External correctness\\filter, micronode, fill=fillviolet3,
            [{\ml{Concise ST}{wang2025concisereasoning}; \ml{LEPA}{zhang2025lepa}; \ml{T$^3$RL}{liao2026t3rl}; \ml{Absolute Zero}{zhao2025absolutezero}}, leafnode, fill=fillviolet4]
          ]
        ]
        [Seed- /\\teacher-supervised, familynode, fill=fillviolet2,
          [Human / cross-model\\supervision, micronode, fill=fillviolet3,
            [{\ml{Self-Instruct}{wang2023selfinstruct}; \ml{Instruction-Backtrans.}{li2024instructionbacktranslation}}, leafnode, fill=fillviolet4]
          ]
        ]
        [External evaluator\\or reward, familynode, fill=fillviolet2,
          [Frozen non-lineage\\scorer, micronode, fill=fillviolet3,
            [{\ml{CSR}{zhou2024csr}}, leafnode, fill=fillviolet4]
          ]
        ]
      ]
    ]
  \end{forest}
\end{lrbox}

\begin{lrbox}{\whenadapttreebox}
  \begin{forest}
    for tree={
      forked edges,
      grow=east,
      reversed=true,
      anchor=center,
      parent anchor=east,
      child anchor=west,
      base=center,
      font=\small,
      rounded corners,
      s sep=1.2pt,
      l sep=4.6pt,
      fork sep=3.2pt,
      inner xsep=1.4pt,
      inner ysep=0.7pt,
      edge+={rounded corners=1.4pt},
    },
    [
      Timing of\\Adaptation,
      rootnode,
      rotate=90,
      anchor=north,
      [Before\\Deployment, tracknode, fill=fillblue1,
        edge={draw=edgeblue, line width=0.9pt, rounded corners=2pt},
        for descendants={edge={draw=edgeblue, line width=0.7pt, rounded corners=1.5pt}},
        [Offline\\Corpus UPT, familynode, fill=fillblue2,
          [{\ml{CPT-LM}{ke2023cptlm}; \ml{Self-Rewarding LM}{yuan2024selfrewarding};
              \ml{R-Zero}{huang2026rzero}; \ml{ScPO}{prasad2025scpo};
              \ml{EM-FT}{agarwal2025unreasonable};
              \ml{VIGOR}{wen2026verifierfreerlllmsintrinsic};
              \ml{V-Zero}{wang2026vzeroselfimprovingmultimodalreasoning};
          \ml{AERO}{gao2026aeroautonomousevolutionaryreasoning}; \ldots}, leafnode, fill=fillblue4]
        ]
      ]
      [Before Target\\Inference, tracknode, fill=fillgreen1,
        edge={draw=edgegreen, line width=0.9pt, rounded corners=2pt},
        for descendants={edge={draw=edgegreen, line width=0.7pt, rounded corners=1.5pt}},
        [Full-Cohort\\Transductive, familynode, fill=fillgreen2,
          [{\ml{TTRL}{zuo2025ttrl}; \ml{MM-UPT}{wei2025mmupt};
              \ml{SCOPE}{wang2025scope}; \ml{COMPASS}{xing2025compass};
          \ml{ECHO}{zhao2026echo}; \ldots}, leafnode, fill=fillgreen4]
        ]
        [Few-Sample\\Target, familynode, fill=fillgreen2,
          [{\ml{TTRV (1/20-sample)}{singh2025ttrv}; \ml{One-shot EM}{gao2025oneshot};
          \ml{RENT}{prabhudesai2025rent}; \ml{RLSC}{li2025rlsc}; \ldots}, leafnode, fill=fillgreen4]
        ]
      ]
      [During Target\\Stream, tracknode, fill=fillviolet1,
        edge={draw=edgeviolet, line width=0.9pt, rounded corners=2pt},
        for descendants={edge={draw=edgeviolet, line width=0.7pt, rounded corners=1.5pt}},
        [Streaming\\Continual, familynode, fill=fillviolet2,
          [{\ml{TLM (online)}{hu2025test}; \ml{TTRV (online)}{singh2025ttrv};
          \ml{SECL}{strich2026secl}; \ml{TT-VLA}{liu2026ttvla}}, leafnode, fill=fillviolet4]
        ]
      ]
      [During Current\\Instance, tracknode, fill=fillorange1,
        edge={draw=edgeorange, line width=0.9pt, rounded corners=2pt},
        for descendants={edge={draw=edgeorange, line width=0.7pt, rounded corners=1.5pt}},
        [Test-Time\\Instance, familynode, fill=fillorange2,
          [{\ml{TTT-NN}{jang2024tttnn}; \ml{SLOT}{hu2025slot};
              \ml{SyTTA}{xu2025you}; \ml{ULDTTA}{xu2026uldtta};
              \ml{Model Whisper}{kang2025modelwhisper};
          \ml{QueST}{song2026queryconditionedtesttimeselftraininglarge}}, leafnode, fill=fillorange4]
        ]
        [Within-\\Sequence, familynode, fill=fillorange2,
          [{\ml{In-Place TTT}{feng2026inplace};
          \ml{PonderTTT}{sim2026ponderttt}; \ldots}, leafnode, fill=fillorange4]
        ]
        [No-Update\\Inference\\(adjacent), familynode, fill=fillviolet2,
          edge={draw=edgeviolet, line width=0.7pt, rounded corners=1.5pt, dashed},
          [{\ml{EM-INF}{agarwal2025unreasonable}; \ldots\,(see\,App.~\ref{app:adjacent})}, leafnode, fill=fillviolet4]
        ]
      ]
    ]
  \end{forest}
\end{lrbox}

\section{Scope, Survey Protocol, and Definitions}
\label{sec:definitions}

\paragraph{Survey protocol.}
We cover text-only and multimodal foundation-model methods from
January~2023 to May~2026 that perform a post-pretraining update on
unlabeled prompts, text, or target inputs. Seed-and-snowball search
covered continued pretraining, test-time adaptation, internal
consensus, self-training, self-rewarding, and multimodal UPT across
ACL Anthology, arXiv, Semantic Scholar, and Google Scholar. The
frozen inventory contains 94 method records from 91 papers: 80 strict
rows from 78 papers, 8 adjacent rows, and 6 prose-only boundary or
antecedent records. A paper can contribute more than one method
record. Appendix~\ref{app:protocol} gives database-specific query
templates, criteria, and counts.

\vspace{-0.25em}
\paragraph{Definition.}
We define \emph{Unsupervised Post-Training} (UPT) as any procedure
that (a)~begins from a finetuned foundation model,
(b)~uses unlabeled prompts, text, or target inputs, (c)~modifies
model parameters, adapters, memories, or persistent local state, and
(d)~computes the update signal without external supervision.
External supervision includes ground-truth answers, verifier
feedback, executable or tool verdicts, human labels, and labels
from a stronger teacher.

\vspace{-0.25em}
\paragraph{Update objects.}
We organize UPT with an \emph{update-object taxonomy}. An internal
update object is the model-derived object that produces the update
signal: a prediction statistic, a relation among model samples,
a self-generated target, or an internal evaluator. The taxonomy is
orthogonal to optimizer, task, modality, and training schedule.

\begin{table*}[t]
  \centering
  \scriptsize
  \setlength{\tabcolsep}{3pt}
  \renewcommand{\arraystretch}{1.05}
  \begin{tabular}{@{}lcccc@{\hskip 6pt}ccccc@{\hskip 6pt}cc@{\hskip 6pt}c@{}}
    \toprule
    & \multicolumn{4}{c}{\textbf{Signal}} & \multicolumn{5}{c}{\textbf{Mechanism}} & \multicolumn{2}{c}{\textbf{Regime}} & \\
    \cmidrule(lr){2-5}\cmidrule(lr){6-10}\cmidrule(lr){11-12}
    \textbf{Method} & NLL & Ent. & Conf. & Geom./Rule & CPT & TTT & EM-min & EM-RL & State & Train & Test & \textbf{LC} \\
    \midrule
    \textsc{CPT-LM}~\citep{ke2023cptlm} & \checkmark & & & & \checkmark & & & & & \checkmark & & \\
    \textsc{Simple CPT}~\citep{qian2024simplescalable} & \checkmark & & & & \checkmark & & & & & \checkmark & & \\
    \textsc{LangAdapt CPT}~\citep{elhady2025languageadapt} & \checkmark & & & & \checkmark & & & & & \checkmark & & \\
    \textsc{Stability-Gap CPT}~\citep{guo2025stabilitygap} & \checkmark & & & & \checkmark & & & & & \checkmark & & \\
    \textsc{ReplayAlign CPT}~\citep{abbes2025replayalign} & \checkmark & & & & \checkmark & & & & & \checkmark & & \\
    \textsc{E2-LLM}~\citep{liu2024e2llm} & \checkmark & & & & \checkmark & & & & & \checkmark & & \checkmark \\
    \textsc{Data Eng 128K}~\citep{fu2024dataengineering} & \checkmark & & & & \checkmark & & & & & \checkmark & & \checkmark \\
    \textsc{LongContext Scaling}~\citep{xiong2024longcontext} & \checkmark & & & & \checkmark & & & & & \checkmark & & \checkmark \\
    \textsc{TLM}~\citep{hu2025test} & \checkmark & & & & & \checkmark & & & & & \checkmark & \\
    \textsc{TTT-NN}~\citep{jang2024tttnn} & \checkmark & & & & & \checkmark & & & & & \checkmark & \\
    \textsc{Long TTT}~\citep{bansal2026qttt} & \checkmark & & & & & \checkmark & & & & & \checkmark & \checkmark \\
    \textsc{In-Place TTT}~\citep{feng2026inplace} & \checkmark & & & & & \checkmark & & & & & \checkmark & \\
    \textsc{EM-FT}~\citep{agarwal2025unreasonable} & & \checkmark & & & & & \checkmark & & & \checkmark & & \\
    \textsc{One-shot EM}~\citep{gao2025oneshot} & & \checkmark & & & & & \checkmark & & & & \checkmark & \\
    \textsc{EM-RL(seq)}~\citep{agarwal2025unreasonable} & & \checkmark & & & & & & \checkmark & & \checkmark & & \\
    \textsc{EM-RL(tok)}~\citep{agarwal2025unreasonable} & & \checkmark & & & & & & \checkmark & & \checkmark & & \\
    \textsc{RENT}~\citep{prabhudesai2025rent} & & \checkmark & & & & & & \checkmark & & \checkmark & & \\
    \textsc{RLSC}~\citep{li2025rlsc} & & & \checkmark & & & & & \checkmark & & \checkmark & & \\
    \textsc{SLOT}~\citep{hu2025slot} & \checkmark & & & & & & & & \checkmark & & \checkmark & \\
    \textsc{SyTTA}~\citep{xu2025you} & & & \checkmark & & & & & & \checkmark & & \checkmark & \\
    \textsc{Model Whisper}~\citep{kang2025modelwhisper} & & & \checkmark & & & & & & \checkmark & & \checkmark & \\
    \textsc{ULDTTA}~\citep{xu2026uldtta} & \checkmark & & & & & & & & \checkmark & & \checkmark & \\
    \textsc{VIGOR}~\citep{wen2026verifierfreerlllmsintrinsic} & & & & \checkmark & & & & \checkmark & & \checkmark & & \\
    \textsc{Latent-GRPO}~\citep{zhang2026silencejudgereinforcementlearning} & & & & \checkmark & & & & \checkmark & & \checkmark & & \\
    \textsc{SSL-R1}~\citep{xie2026sslr1selfsupervisedvisualreinforcement} & & & & \checkmark & & & & \checkmark & & \checkmark & & \\
    \textsc{SUDER}$^{\ddagger}$~\citep{hong2025suderselfimprovingunifiedlarge} & & & & \checkmark & & & & \checkmark & & \checkmark & & \\
    \bottomrule
  \end{tabular}
  \vspace{-0.5em}
  \caption{Strict UPT methods in Family~I (\S\ref{sec:family-i}).
    \emph{Signal}: NLL, entropy (Ent.), self-confidence (Conf.), or
    other geometric/rule-based statistic (Geom./Rule).
    \emph{Mechanism}: continued pretraining (CPT), test-time training
    (TTT), entropy/confidence minimization (EM-min), entropy/confidence
    or other internal-statistic policy-gradient reward (EM-RL),
    sample-local state update (State).
    \emph{LC}: long-context target.
    $^{\ddagger}$ Family~I/IV bridge case (\S\ref{sec:family-iv}).
  }
  \vspace{-1em}
  \label{tab:family-i}
\end{table*}

\vspace{-0.25em}
\paragraph{Boundary checks and update-object rule.}
\label{sec:boundary-protocol}
A method is \emph{strict UPT} if all four checks hold:
\begin{itemize}\setlength\itemsep{0pt}\setlength\topsep{2pt}
  \item[\textbf{B1.}] It performs an explicit update to parameters,
    adapters, memories, or persistent local state.
  \item[\textbf{B2.}] The update signal is only from
    unlabeled inputs \& same-lineage samples or judgments.
  \item[\textbf{B3.}] No external supervision enters the update.
  \item[\textbf{B4.}] Any judge, scorer, or reward model used in the
    update derives from the same model lineage.
\end{itemize}
Family assignment follows the object consumed by the gradient. A
multi-sample majority statistic used as a reward belongs
to Sample-Relation Supervision, whereas one used to build
pseudo-labels, curricula, or preference pairs belongs to
Self-Generated Target Bootstrapping. The four checks rule out
\emph{explicit} external supervision; the separate audit below
records structural properties of the task and evaluation protocol.

\vspace{-0.25em}
\paragraph{Task-structure audit.}
We separately record whether a result relies on an answer extractor
or canonicalizer, a finite answer alphabet, a code signature,
full-cohort transductive access, or an open-ended output space.
These features define equivalence classes that make model samples
easier to group and compare; correctness-bearing executions are
treated as external verdicts under~\bcref{B3}.
Table~\ref{tab:task-prior-audit} separates these structural priors
from correctness-bearing supervision in the representative-evidence
audit presented in Appendix~\ref{app:synthesis-evidence}.

\vspace{-0.25em}
\paragraph{Adjacent methods.}
The adjacent track retains five neighboring paradigms for comparison:
no-update inference-time optimization; verifier- or tool-assisted
self-training; human- or seed-supervised bootstrapping;
stronger-teacher or cross-model distillation; and external reward
or evaluator
methods~\citep{liao2026t3rl,zhao2025absolutezero,wang2023selfinstruct,li2024instructionbacktranslation}.
This parallel organization preserves a precise internal-signal core
while retaining the broader design space.

\vspace{-0.25em}
\paragraph{Formal setup.}
Let $f_\theta$ be a foundation model and let $\mathcal{D}_x$ be a
distribution of unlabeled prompts, text, or target inputs.
$\theta$ denotes the updatable state, including
parameters, adapters, memories, or persistent local state. A UPT
procedure has three components. First, a sampling or aggregation
operator $\mathcal{S}$ produces an internal update object
$z = \mathcal{S}(f_\theta, x), x \sim \mathcal{D}_x,$
such as predictive distribution, a set of rollouts, candidate
answers, preference pairs, generated targets, or judge verdicts.
Second, a signal extractor $\sigma$ maps this object to an update
signal
$s = \sigma(z),$
which may be scalar, pairwise, token or sequence-level. Third,
an update operator $\mathcal{U}$ maps these to an updated state:
\[
  \theta' = \mathcal{U}(\theta; x, z, s).
\]
We call $z$, not $\sigma\!\circ\!\mathcal{S}$, the
\emph{internal update object}. The update-object taxonomy classifies a
method by the type of $z$ used to compute the update signal.

\begin{table*}[t]
  \centering
  \scriptsize
  \setlength{\tabcolsep}{4pt}
  \renewcommand{\arraystretch}{1.05}
  \begin{tabular}{@{}lcccccc@{\hskip 6pt}cc@{\hskip 6pt}c@{}}
    \toprule
    & \multicolumn{6}{c}{\textbf{Consensus signal}} & \multicolumn{2}{c}{\textbf{Regime}} & \\
    \cmidrule(lr){2-7}\cmidrule(lr){8-9}
    \multirow{2}{*}{\textbf{Method}}
    & Majority & Semantic & Self- & Path & Pairwise & Soft
    & Train & Test
    & \multirow{2}{*}{\textbf{\quad Multimodal}} \\
    & vote & cluster & certainty & consist. & agreement & cluster & & & \\
    \midrule
    \textsc{EMPO}~\citep{zhang2025rightquestion} & & \checkmark & & & & & \checkmark & & \\
    \textsc{Intuitor}~\citep{zhao2025intuitor} & & & \checkmark & & & & \checkmark & & \\
    \textsc{CoVo}~\citep{zhang2025covo} & & & & \checkmark & & & \checkmark & & \\
    \textsc{Co-rewarding}~\citep{zhang2026corewarding} & & & & & \checkmark & & \checkmark & & \\
    \textsc{EvoLMM}~\citep{thawakar2025evolmm} & \checkmark & & & & & & \checkmark & & \checkmark \\
    \textsc{TTRL}~\citep{zuo2025ttrl} & \checkmark & & & & & & & \checkmark & \\
    \textsc{ETTRL}~\citep{liu2025ettrl} & \checkmark & & & & & & & \checkmark & \\
    \textsc{ECHO}~\citep{zhao2026echo} & \checkmark$^\dagger$ & & & & & & & \checkmark & \\
    \textsc{SPINE}~\citep{wu2025spine} & \checkmark$^\dagger$ & & & & & & & \checkmark & \\
    \textsc{Self-Harmony}~\citep{wang2026selfharmony} & & & & & \checkmark & & & \checkmark & \\
    \textsc{DARE}~\citep{du2026dare} & & & & & & \checkmark & & \checkmark & \\
    \textsc{SCOPE}~\citep{wang2025scope} & \checkmark & & & & & & \checkmark & & \\
    \textsc{COMPASS}~\citep{xing2025compass} & \checkmark & & & & & & \checkmark & & \\
    \textsc{SCRL}~\citep{yan2026scrl} & \checkmark & & & & & & \checkmark & & \\
    \textsc{RLCCF}~\citep{yuan2025rlccf} & & & & & \checkmark & & \checkmark & & \\
    \textsc{RoiRL}~\citep{arzhantsev2025roirl} & \checkmark & & & & & & \checkmark & & \\
    \textsc{EVOL-RL}~\citep{zhou2025evolrl} & \checkmark & & & & & & \checkmark & & \\
    \textsc{TTRV}~\citep{singh2025ttrv} & \checkmark & & & & & & & \checkmark & \checkmark \\
    \textsc{MM-UPT}~\citep{wei2025mmupt} & \checkmark & & & & & & \checkmark & & \checkmark \\
    \textsc{Dual~Consensus}~\citep{du2026dualconsensusescapingspurious} & \checkmark & & & & & & \checkmark & & \\
    \textsc{CSRS}~\citep{yu2026stabilizingunsupervisedselfevolutionmllms} & & & & & & \checkmark & \checkmark & & \checkmark \\
    \textsc{EvoQuality}~\citep{wen2026selfevolvingvisionlanguagemodelsimage} & & & & & \checkmark & & \checkmark & & \checkmark \\
    \bottomrule
  \end{tabular}
  \vspace{-0.5em}
  \caption{Strict UPT methods in Family~II (\S\ref{sec:family-ii}).
    \emph{Consensus signal}: which multi-sample statistic over
    multiple internal samples drives the gradient.
    $^\dagger$ marks methods where the consensus is the
    \emph{selection} mask and a separate intrinsic term shapes
  advantages or tokens.}
  \vspace{-1.5em}
  \label{tab:family-ii}
\end{table*}

\vspace{-0.25em}
\section{Prediction-Statistic Optimization}
\vspace{-0.25em}
\label{sec:family-i}

The first family is direct: $\sigma$ reads a scalar from
$f_\theta$ at a single observation, and $\mathcal{U}$ optimizes
that scalar. This makes it the historical baseline, with continued
pretraining and test-time training predating the other three
families. By construction, it generates no pseudo-labels,
constructs no preference pairs, and trains no internal judge: the
update object is the predictive quantity (token NLL,
sequence likelihood, entropy, confidence) or an analogous
geometric or rule-based statistic. Table~\ref{tab:family-i}
catalogs 26 strict UPT methods along
Signal\,$\times$\,Mechanism.

\vspace{-0.5em}
\paragraph{Predictive likelihood minimization.}
The basic instantiation is CPT:
minimizing LM loss on unlabeled text. CPT
recipes~\citep{ke2023cptlm,qian2024simplescalable} need no
annotation when domain shift is the bottleneck; refinements address
failure
modes~\citep{guo2025stabilitygap,abbes2025replayalign,elhady2025languageadapt},
and a parallel sub-line scales context windows by corpus engineering
alone~\citep{liu2024e2llm,fu2024dataengineering,xiong2024longcontext}.
The NLL objective also reaches into TTA:
\textsc{TLM}~\citep{hu2025test} treats the test stream as a
CPT corpus; \textsc{TTT-NN}~\citep{jang2024tttnn} restricts each
update to a retrieved neighborhood;
\textsc{Long~TTT}~\citep{bansal2026qttt} extends it to long
context; \textsc{In-Place~TTT}~\citep{feng2026inplace}
interleaves updates with generation.

\vspace{-0.25em}
\paragraph{Entropy and confidence objectives.}
A second sub-line generalizes the read-out to \emph{predictive
confidence}, exploiting that high-quality reasoning traces are
low-entropy. Entropy
minimization~\citep{agarwal2025unreasonable,gao2025oneshot} treats
per-token entropy as a training loss or single-prompt gradient.
The statistics wrap into policy-optimization loops:
\textsc{EM-RL}~\citep{agarwal2025unreasonable} recasts entropy as
intrinsic reward in GRPO;
\textsc{RENT}~\citep{prabhudesai2025rent} and
\textsc{RLSC}~\citep{li2025rlsc} reward self-confidence with no
answer key. \citet{zhang2025nofreelunch} show that initialization
and update duration shape entropy- and confidence-based
optimization, motivating stratified evaluation in
\S\ref{sec:open-problems}.

\vspace{-0.25em}
\paragraph{Beyond entropy and confidence.}
A recent line broadens Family~I to richer internal statistics
(\emph{Geom./Rule} column of Table~\ref{tab:family-i}).
\textsc{VIGOR}~\citep{wen2026verifierfreerlllmsintrinsic} uses
policy's teacher-forced gradient norm as intrinsic GRPO reward;
\textsc{Latent-GRPO}~\citep{zhang2026silencejudgereinforcementlearning}
replaces external judges with terminal hidden-state geometry. In
multimodal settings,
\textsc{SSL-R1}~\citep{xie2026sslr1selfsupervisedvisualreinforcement}
derives rewards from visual self-supervised puzzles, and
\textsc{SUDER}~\citep{hong2025suderselfimprovingunifiedlarge} uses
reverse-task likelihood as a prediction statistic bridging
Families~I and~IV (\S\ref{sec:family-iv}).

\vspace{-0.25em}
\paragraph{Sample-local state update.}
A third sub-line moves the update target from full parameters to a
small, sample-local state: a per-prompt optimization
vector~\citep{hu2025slot}, a test-time LoRA~\citep{xu2025you}, a
steering vector~\citep{kang2025modelwhisper}, or layer-wise dynamic
adaptation~\citep{xu2026uldtta}. These methods optimize the same
prediction-statistic objectives but with a smaller update footprint
and shorter persistence; we trace this timing axis explicitly in
\S\ref{sec:timing}.

\begin{table*}[t]
  \centering
  \scriptsize
  \setlength{\tabcolsep}{2pt}
  \renewcommand{\arraystretch}{1.05}
  \begin{tabular}{@{}lcccc@{\hskip 4pt}cccccccccc@{\hskip 4pt}cc@{}}
    \toprule
    & \multicolumn{4}{c}{\textbf{Generated target}}
    & \multicolumn{10}{c}{\textbf{Selection criterion}}
    & \multicolumn{2}{c}{\textbf{Regime}} \\
    \cmidrule(lr){2-5}\cmidrule(lr){6-15}\cmidrule(lr){16-17}
    \textbf{Method} & Instr. & Rationale & Curric. & Pref. %
    & Doc. & KB & Cycle & SC & Confid. & Maj. & Debate & MBR & Foresight & Reflect %
    & Train & Test \\
    \midrule
    \textsc{Self-Tuning}$^\dagger$~\citep{zhang2025selftuning} & \checkmark & & & & \checkmark & & & & & & & & & & \checkmark & \\
    \textsc{KBAlign}~\citep{zeng2025kbalign} & \checkmark & & & & & \checkmark & & & & & & & & & \checkmark & \\
    \textsc{CYCLE-INSTRUCT}$^\dagger$~\citep{shen2025cycleinstruct} & \checkmark & & & & & & \checkmark & & & & & & & & \checkmark & \\
    \textsc{Self-Improve}~\citep{huang2023selfimprove} & & \checkmark & & & & & & \checkmark & & & & & & & \checkmark & \\
    \textsc{Quiet-STaR}~\citep{zelikman2024quietstar} & & \checkmark & & & & & & & & & & & & & \checkmark & \\
    \textsc{Confident ST}~\citep{wang2025confidentreasoning} & & \checkmark & & & & & & & \checkmark & & & & & & \checkmark & \\
    \textsc{GENIUS}~\citep{xu2025genius} & & \checkmark & & & & & & & & & & & \checkmark & & \checkmark & \\
    \textsc{LRM Self-Train}~\citep{shi2025lrmselftrain} & & \checkmark & & & & & & & & \checkmark & & & & & \checkmark & \\
    \textsc{DTE}~\citep{liu2025dte} & & \checkmark & & & & & & & & & \checkmark & & & & \checkmark & \\
    \textsc{LongMagpie}~\citep{gao2025longmagpie} & & \checkmark & & & & & & & & & & & & & \checkmark & \\
    \textsc{Long Self-Improve}~\citep{wang2024longselfimprove} & & \checkmark & & & & & & & & & & \checkmark & & & \checkmark & \\
    \textsc{TTCS}~\citep{yang2026ttcs} & & & \checkmark & & & & & \checkmark & & & & & & & & \checkmark \\
    \textsc{DiSCTT}~\citep{moradi2026disctt} & & & \checkmark & & & & & \checkmark & & & & & & & & \checkmark \\
    \textsc{TTSR}~\citep{he2026ttsr} & & & \checkmark & & & & & & & & & & & \checkmark & & \checkmark \\
    \textsc{R-Zero}~\citep{huang2026rzero} & & & \checkmark & & & & & & & \checkmark & & & & & \checkmark & \\
    \textsc{ScPO}~\citep{prasad2025scpo} & & & & \checkmark & & & & \checkmark & & & & & & & \checkmark & \\
    \textsc{MACA}~\citep{samanta2025maca} & & & & \checkmark & & & & & & & \checkmark & & & & \checkmark & \\
    \textsc{LongPO}~\citep{chen2025longpo} & & & & \checkmark & & & & & & & & & & & \checkmark & \\
    \textsc{RLSF}~\citep{vanniekerk2025rlsf} & & & & \checkmark & & & & & \checkmark & & & & & & \checkmark & \\
    \textsc{G-Zero}~\citep{huang2026gzeroselfplayopenendedgeneration} & & & & \checkmark & & & & & \checkmark & & & & & & \checkmark & \\
    \textsc{QueST}~\citep{song2026queryconditionedtesttimeselftraininglarge} & \checkmark & & & & & & & \checkmark & & & & & & & & \checkmark \\
    \textsc{V-Zero}~\citep{wang2026vzeroselfimprovingmultimodalreasoning} & \checkmark & & & & & & & & & \checkmark & & & & & \checkmark & \\
    \bottomrule
  \end{tabular}
  \vspace{-0.5em}
  \caption{Strict UPT methods in Family~III (\S\ref{sec:family-iii}).
    \emph{Generated target}: instruction/response (Instr.),
    rationale, curriculum (Curric.), or preference pair (Pref.).
    \emph{Selection criterion}: document grounding (Doc.),
    knowledge-base grounding (KB), cycle consistency, self-consistency
    (SC), internal confidence (Confid.), majority vote (Maj.),
    multi-agent debate, minimum-Bayes risk (MBR), stepwise foresight
    (Foresight), self-reflection (Reflect).
    $^\dagger$ marks a stage- or variant-qualified strict entry.
  }
  \vspace{-1em}
  \label{tab:family-iii}
\end{table*}

\vspace{-0.25em}
\section{Sample-Relation Supervision}
\vspace{-0.25em}
\label{sec:family-ii}

The second family changes what $\sigma$ computes. Methods in
Family~I read a scalar from the model's distribution at a single
observation; Family~II reads a \emph{relation} across multiple
internal update objects, such as rollouts, paraphrases, candidate
answers, or multiple agents. The internal update object is a
\emph{multi-sample statistic}: a cluster mass, a consistency score, a vote, or a
contrastive agreement. Table~\ref{tab:family-ii} shows that 13/22
methods reduce the relation to a binary majority vote;
semantic-cluster, self-certainty, pairwise-agreement, and
softened-frequency variants populate the long tail.

\vspace{-0.25em}
\subsection{Self-Consistency Within a Single Prompt}
\vspace{-0.25em}
\label{sec:f2-self}

The simplest relation operates among multiple samples drawn from
the same prompt. \textsc{EMPO}~\citep{zhang2025rightquestion}
clusters rollouts and uses cluster mass or semantic entropy as the
reward. \textsc{Intuitor}~\citep{zhao2025intuitor} replaces external
rewards with \emph{self-certainty}, the forward KL to a uniform
distribution. \textsc{CoVo}~\citep{zhang2025covo} combines path
consistency and volatility, while
\textsc{Co-rewarding}~\citep{zhang2026corewarding} co-evolves a pair
of networks and reads their contrastive agreement as the reward,
which sidesteps the reward hacking that arises when one network
grades itself.

\vspace{-0.25em}
\subsection{Consensus and Test-Time RL}
\vspace{-0.25em}
\label{sec:f2-consensus}

A larger sub-line scales the relation across many samples and feeds
it back into a policy-optimization loop. The canonical instance is
\textsc{TTRL}~\citep{zuo2025ttrl} (Table~\ref{tab:family-ii},
majority-vote rows): for each test prompt, it draws $N$ rollouts,
takes the majority answer as a pseudo-label, and runs a GRPO step
against it. Refinements form a coherent line: entropy-regularized
exploration~\citep{liu2025ettrl}; paraphrase consistency and
self-play~\citep{wang2026selfharmony}; soft rewards from rollout
statistics~\citep{du2026dare}; consensus across model
populations~\citep{yuan2025rlccf}; entropy-shaped
advantages~\citep{zhao2026echo}; and entropy-band token
masks~\citep{wu2025spine}. The same recipe also runs at training
time on unlabeled
prompts~\citep{wang2025scope,xing2025compass,yan2026scrl}.
\textsc{EVOL-RL}~\citep{zhou2025evolrl} isolates semantic novelty as
a mechanism for preserving rollout diversity under majority-based
optimization, exposing a broader diversity--selection trade-off for
sample-relation methods.

Follow-up works generalize the relation beyond a single hard-majority
signal. \textsc{Dual~Consensus}~\citep{du2026dualconsensusescapingspurious}
uses an anchor-explorer voting relation to escape spurious
majorities; \textsc{CSRS}~\citep{yu2026stabilizingunsupervisedselfevolutionmllms}
softens hard majority rewards into frequency signals over retraced
multimodal reasoning sets, while
\textsc{EvoQuality}~\citep{wen2026selfevolvingvisionlanguagemodelsimage}
turns VLM pairwise judgments into majority-voted
image-quality pseudo-rankings.

\vspace{-0.25em}
\subsection{Cross-Modal Extension}
\vspace{-0.25em}
\label{sec:f2-multimodal}

The relational recipe transfers to multimodal foundation models
through modality-specific reductions of free-form outputs.
\textsc{TTRV}~\citep{singh2025ttrv} ports TTRL to vision-language
models with a frequency-plus-entropy reward; on ImageNet,
InternVL3-8B reaches 99.31\% (vs.\ GPT-4o's 98.30\%) and exceeds
GPT-4o by 2.3 points across eight benchmarks.
\textsc{MM-UPT}~\citep{wei2025mmupt} runs a training-time variant
on Qwen2.5-VL with self-generated prompts;
\textsc{EvoLMM}~\citep{thawakar2025evolmm} closes a
proposer--solver loop driven by multi-rollout consistency. All
three derive visual supervision from the model's own rollouts,
without image-text labels, captioners, or external verifiers.

\vspace{-0.25em}
\section{Self-Generated Target Bootstrapping}
\vspace{-0.25em}
\label{sec:family-iii}

The third family \emph{constructs} a trainable target from the model
distribution, such as an instruction, a rationale, a plan, a debate
trace, a curriculum, or a preference pair. The update operator
$\mathcal{U}$ then applies a standard SFT or DPO step against that
object. We organize the family by the type of object that the model
bootstraps
(\S\ref{sec:f3-instruction}--\S\ref{sec:f3-pref}).
Table~\ref{tab:family-iii} shows that rationales form the largest branch
with 8 methods, followed by instructions and preferences with 5
each and curricula with 4.

\vspace{-0.25em}
\subsection{Self-Curated Instructions and Knowledge}
\vspace{-0.25em}
\label{sec:f3-instruction}

The simplest synthetic targets are prompt--response pairs derived
from raw documents. \textsc{Self-Tuning}~\citep{zhang2025selftuning}
turns documents into a staged memorization--comprehension--reflection
curriculum: next-token prediction, automatically constructed
document tasks, and closed-book reconstruction. Its document-derived
self-teaching component falls within the strict core.

\textsc{KBAlign}~\citep{zeng2025kbalign} self-annotates short- and
long-dependency question--answer pairs from a textual knowledge
base, tunes on them, and then uses its own predictions and generated
correction rationales in later rounds. \textsc{CYCLE-INSTRUCT}~\citep{shen2025cycleinstruct}
instead instantiates two transformer models from the same base: a
forward model $M_{Q\rightarrow A}$ and a backward model
$M_{A\rightarrow Q}$. They alternate pseudo-answer generation,
backward reconstruction training, pseudo-instruction generation,
and forward reconstruction training. The shared-base dual loop
anchors both reconstruction directions to the original text and
limits drift across successive rounds of self-training.

\vspace{-0.25em}
\subsection{Self-Trained Rationales and Curricula}
\vspace{-0.25em}
\label{sec:f3-rationale}

A larger sub-line bootstraps reasoning targets. The seminal recipe
is \textsc{Self-Improve}~\citep{huang2023selfimprove}
(Table~\ref{tab:family-iii}, rationale rows): generate multiple CoT
traces, filter them by self-consistency, and fine-tune on the
survivors. Variants extend the recipe to latent thoughts during
CPT~\citep{zelikman2024quietstar}, confidence-selected
traces~\citep{wang2025confidentreasoning}, foresight-resampled
sequences~\citep{xu2025genius}, multi-agent debate
trajectories~\citep{liu2025dte}, long-context
extensions~\citep{gao2025longmagpie,wang2024longselfimprove}, and
\textsc{R-Zero}~\citep{huang2026rzero}, which splits a base model
into a Challenger (proposing problems at the Solver's ability
boundary) and a Solver (training on majority-vote pseudo-labels);
both the curriculum and the labels are internally synthesized. A
test-time variant produces curricula at
inference~\citep{yang2026ttcs,moradi2026disctt,he2026ttsr}: the
machinery uses consensus (Family~II in isolation), but the
gradient is computed against the synthesized target, so the
update-object rule places it in Family~III.
\textsc{LRM~Self-Train}~\citep{shi2025lrmselftrain} follows the
same logic: its reward is a binary majority match, but the stated
unit of analysis is the pseudo-target that evolves alongside the
solver.

A recent line extends self-bootstrapping beyond math curricula.
\textsc{G-Zero}~\citep{huang2026gzeroselfplayopenendedgeneration}
runs a proposer--generator loop creating hint-driven preference
pairs for open-ended generation without any external judge.
\textsc{QueST}~\citep{song2026queryconditionedtesttimeselftraininglarge}
moves the loop to test time, generating query-conditioned
auxiliary problems and fitting a small LoRA adapter before answering;
\textsc{V-Zero}~\citep{wang2026vzeroselfimprovingmultimodalreasoning}
trains a questioner--solver vision-language loop on unlabeled
images with its own questions and majority-vote pseudo-labels.

\vspace{-0.25em}
\subsection{Self-Generated Preference Pairs}
\vspace{-0.25em}
\label{sec:f3-pref}

A final sub-line synthesizes preference pairs and optimizes with
DPO-style objectives: ranking responses by
self-consistency~\citep{prasad2025scpo}; multi-agent debate
consensus~\citep{samanta2025maca}; \emph{short-to-long} pairs (same
instruction on short vs.\ long context) for long-context
self-evolution~\citep{chen2025longpo}; and ranking chains of thought
by an internal answer-confidence statistic~\citep{vanniekerk2025rlsf}.
\textsc{Confident ST} and \textsc{RLSF} select SFT or DPO targets
by confidence, so the update-object rule assigns both to
Family~III; Appendix~\ref{app:reliability} contrasts
evaluator-reward variants.

\vspace{-0.25em}
\section{Internal Evaluator Bootstrapping}
\vspace{-0.25em}
\label{sec:family-iv}

\begin{table}[t]
  \centering
  \scriptsize
  \setlength{\tabcolsep}{3pt}
  \renewcommand{\arraystretch}{1.05}
  \begin{tabularx}{\linewidth}{@{}lcL@{}}
    \toprule
    \textbf{Method} & \textbf{Update} & \textbf{Issue addressed} \\
    \midrule
    \textsc{Self-Rewarding LM}~\citep{yuan2024selfrewarding} & DPO & baseline self-judge \\
    \textsc{CREAM}~\citep{wang2024cream} & DPO & bias amplification \\
    \textsc{Meta-Rewarding}~\citep{wu2024metarewarding} & DPO & judge saturation \\
    \textsc{Temporal SRLM}~\citep{jin2025temporalselfrewarding} & DPO & gradient vanishing \\
    \textsc{CoNL}~\citep{sui2026conl} & PG & critique-as-reward \\
    \textsc{RLME}~\citep{rentschler2026rlme} & PG & NL meta-judge \\
    \textsc{Meta-TTRL}~\citep{tan2026metattrl} & PG & test-time MM rubric \\
    \textsc{AERO}~\citep{gao2026aeroautonomousevolutionaryreasoning} & PG & endogenous critique \\
    \textsc{Self-Judge}~\citep{wu2026modelsjudgethemselvesunsupervised} & PG & judge-gated reward \\
    \textsc{GvU}$^{\ddagger}$~\citep{pan2026learninggenerateunderstandingunderstandingdriven} & PG & cross-branch evaluator \\
    \bottomrule
  \end{tabularx}
  \vspace{-0.5em}
  \caption{Strict UPT methods in Family~IV (\S\ref{sec:family-iv}).
  $^{\ddagger}$ Family~I/IV bridge case (discussed below).}
  \vspace{-1.5em}
  \label{tab:family-iv}
\end{table}

In Family~IV, the model also self-elevates the \emph{judge}: the
internal update object is an evaluator (scorer, reward model, or
meta-judge) produced and consumed by the same lineage. Where
Family~III bootstraps a \emph{trainable target} and trains against
it, Family~IV bootstraps a \emph{trainable verdict-emitter} and
trains its outputs through this verdict. Because the verdict can
take two natural forms, a pairwise preference or a scalar score,
Table~\ref{tab:family-iv} splits the 10 methods: 4
DPO~variants consume preference pairs from the judge, and 6
policy-gradient variants treat the judge's score as reward. Both
sub-lines share one constraint: the evaluator must come from the same
model lineage; otherwise Appendix~\ref{app:adjacent} classifies the
method as adjacent under~\bcref{B4}.

\vspace{-0.25em}
\subsection{Self-Rewarding via Internal Judges}
\vspace{-0.25em}
\label{sec:f4-judge}

The line opens with \textsc{Self-Rewarding~LM}~\citep{yuan2024selfrewarding}:
one LLM alternates between actor and judge roles; chosen--rejected
pairs feed iterative DPO, and both capabilities improve in tandem.
Subsequent methods stabilize the self-rewarding loop along three
dimensions: \textsc{CREAM} enforces cross-iteration
consistency~\citep{wang2024cream}, \textsc{Meta-Rewarding}
introduces meta-judgment~\citep{wu2024metarewarding}, and
\textsc{Temporal SRLM} anchors preferences across model
generations~\citep{jin2025temporalselfrewarding}.
\textsc{CSR}~\citep{zhou2024csr} extends self-rewarding to
vision-language models with a calibrated self-judge;
Appendix~\ref{app:adjacent} audits its external CLIP reward term.

\vspace{-0.25em}
\subsection{Evaluator-Driven Policy Optimization}
\vspace{-0.25em}
\label{sec:f4-evalrl}

A second sub-line lets the evaluator drive a policy-gradient update
directly: structured multi-agent debate with critique-helpfulness
as Bradley--Terry-aggregated reward~\citep{sui2026conl}; an internal
evaluator's natural-language meta-judgments (``\textit{correct? / logically
consistent?}'') as scalar rewards~\citep{rentschler2026rlme}; and a
multimodal cohort-visible test-time recipe via a metacognitive
introspector~\citep{tan2026metattrl}. Family~IV routes reward through
an explicit evaluator, whereas Family~II derives it from
multi-sample statistics, a distinction documented in
Appendix~\ref{app:reliability}.

The same internal-evaluator pattern appears in broader
self-evolution loops.
\textsc{AERO}~\citep{gao2026aeroautonomousevolutionaryreasoning}
uses self-generated tasks and counterfactual criticism for
KTO-style updates, and
\textsc{Self-Judge}~\citep{wu2026modelsjudgethemselvesunsupervised}
uses a same-model-lineage frozen judge to modulate actor self-consistency.
Two unified multimodal systems sit on the Family~I/IV boundary:
the update-object rule assigns
\textsc{SUDER}~\citep{hong2025suderselfimprovingunifiedlarge} to
Family~I and
\textsc{GvU}~\citep{pan2026learninggenerateunderstandingunderstandingdriven}
to Family~IV according to the primary signal each update consumes.

\vspace{-0.25em}
\section{Cross-Family Synthesis}
\vspace{-0.25em}
\label{sec:synthesis}

\vspace{-0.25em}
\paragraph{Four sources of leverage.}
UPT converts an internal proxy into an update: Family~I exploits
predictive statistics; Family~II aggregates evidence across samples;
Family~III builds inspectable targets; and Family~IV supplies semantic
criteria for open-ended outputs. Increasing semantic flexibility also
lengthens the feedback path, making consistency, diversity, and
independent evaluation increasingly valuable.

Self-training theory links improvement from unlabeled data to
class-consistent neighborhoods and
expansion~\citep{wei2021theoreticalselftraining}, while evidence on
confirmation bias motivates consistency, diversity, and held-out
checks~\citep{arazo2020pseudolabeling}.
Appendix~\ref{app:synthesis-evidence} maps one representative result
per family and signal-shaping task structures to these conditions
(Tables~\ref{tab:representative-evidence} and
\ref{tab:task-prior-audit}).

\vspace{-0.25em}
\paragraph{From signal to update.}
The four families differ in where uncertainty is converted into a
training decision. Family~I optimizes a statistic at one model state.
\textsc{LangAdapt CPT}~\citep{elhady2025languageadapt} uses token
likelihood on domain text, while entropy- and confidence-based variants
reshape the predictive distribution; this short path depends on
alignment between the statistic and downstream behavior rather than
proxy sharpness alone. Family~II aggregates rollouts.
\textsc{TTRL}~\citep{zuo2025ttrl} canonicalizes sampled answers, turns
majority-class membership into a GRPO reward, and updates the policy
against that relation. Agreement is informative when errors are diverse
and answer equivalence is stable, but self-reinforcing when errors
correlate. Family~III materializes selected outputs as inspectable,
reusable targets. In
\textsc{R-Zero}~\citep{huang2026rzero}, a Challenger proposes problems
near the Solver's ability boundary and the Solver trains on synthesized
labels. The resulting data can be inspected and reused, while selection
errors can persist after entering the target set. Family~IV inserts a
same-lineage evaluator before the update.
\textsc{Self-Rewarding LM}~\citep{yuan2024selfrewarding} and
\textsc{CoNL}~\citep{sui2026conl} turn judgments or critiques into
rewards for open-ended outputs, while actor--judge drift becomes part
of the dynamics. Semantic reach grows with aggregation,
target construction, and evaluation, together with the longer
feedback path through which errors can recur.

\vspace{-0.25em}
\paragraph{Same observable, different gradient path.}
Majority vote and confidence recur across families, so classification
follows the object consumed by the gradient. \textsc{TTRL} and
\textsc{RoiRL}~\citep{arzhantsev2025roirl} consume majority agreement
as reward, placing them in Family~II; \textsc{LRM
Self-Train}~\citep{shi2025lrmselftrain} uses it to select solutions and
computes an SFT loss against the retained target set, placing it in
Family~III. \textsc{Confident
ST}~\citep{wang2025confidentreasoning} and
\textsc{RLSF}~\citep{vanniekerk2025rlsf} follow the same target path
when confidence selects trajectories before SFT or DPO. Confidence
consumed directly as loss or reward is a Family~I statistic; a
same-lineage evaluator's semantic score consumed as reward is a
Family~IV object. This gradient-path rule remains stable across labels
such as self-training, self-rewarding, and confidence optimization;
Appendix~\ref{app:reliability} records these assignments and the update
objects that justify them across the taxonomy.

\vspace{-0.25em}
\paragraph{Reading reported evidence.}
Table~\ref{tab:representative-evidence} preserves one
mechanism-matched result per family under its original setup.
For Family~I, \textsc{LangAdapt CPT} reduces Basque perplexity from
23.64 to 3.35 and raises aggregate downstream accuracy from 27.43 to
34.14, separating corpus adaptation from task transfer. For Family~II,
\textsc{TTRL} raises AIME~2024 from 12.9 to 40.2 and MATH-500 from
46.7 to 83.4, while its GPQA result ties majority reward to the target
task structure. For Family~III,
\textsc{Quiet-STaR}~\citep{zelikman2024quietstar} raises zero-shot
GSM8K from 5.9 to 10.9 and CommonsenseQA from 36.3 to 47.2,
supporting future-token selection of latent reasoning targets. For
Family~IV, \textsc{CoNL} raises AIME~2024 from 60.0 to 76.5 and
DeepMath from 70.5 to 87.1, demonstrating a strong same-policy
semantic signal. Together, the rows connect the four update objects
to corpus fit, answer equivalence, target selection, and evaluator
quality; Appendix~\ref{app:synthesis-evidence} retains the
source-specific metrics, budgets, and adaptation protocols.

\vspace{-0.25em}
\paragraph{Choosing by available signal.}
When adaptation data are raw documents and distribution shift is
the main problem, Family~I is the direct baseline. When multiple
samples are affordable and outcomes admit a defensible equivalence
relation, Family~II can exploit agreement. When generated targets
can be inspected before an offline update, Family~III offers the
clearest data interface. When outputs are open-ended and equality is
undefined, Family~IV supplies learned semantic criteria through
evaluator-mediated rollouts; externally grounded programmatic
verifiers remain adjacent.

\vspace{-0.25em}
\paragraph{Safeguards follow the feedback path.}
Family~I safeguards track calibration and downstream quality
alongside the optimized statistic, distinguishing a sharper proxy from
a better model. Family~II tracks sample diversity, wrong-majority
frequency, and agreement under alternative canonicalizers. Family~III
tracks target diversity and refreshes the generated dataset before
selection errors persist across rounds. Family~IV tracks actor--judge
correlation, preference margins, cross-round consistency, and
saturation. Across families, frozen baselines, rollback checkpoints,
and held-out evaluation test whether gains persist. Each check acts
where the internal proxy becomes a training signal, interrupting error
accumulation before the next round reads a more biased proxy.

\vspace{-0.25em}
\section{Timing of Adaptation}
\vspace{-0.25em}
\label{sec:timing}

\begin{figure*}[!t]
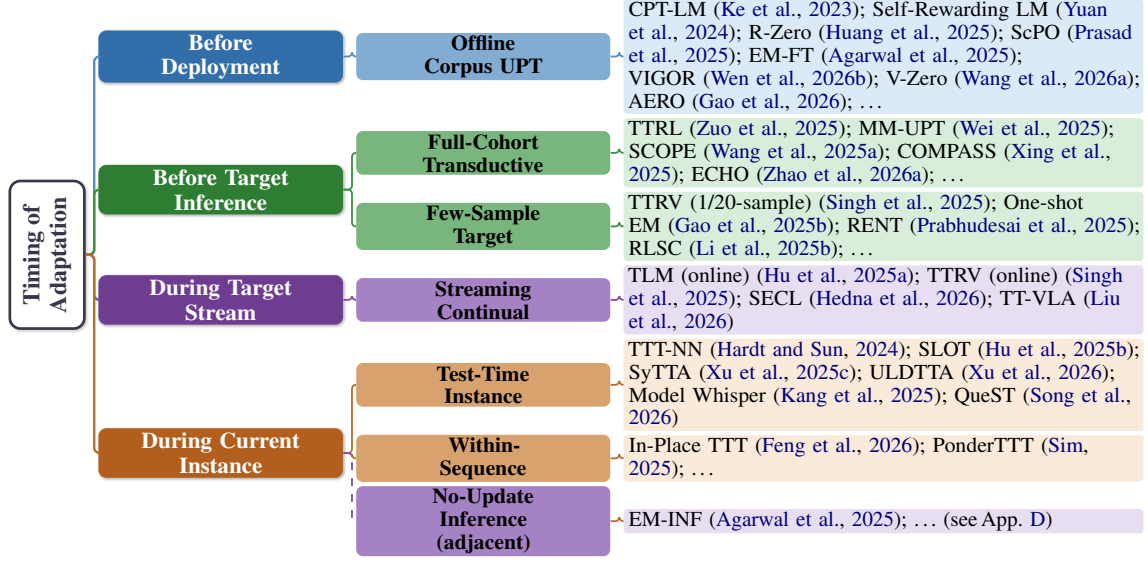

  \centering
  \begin{adjustbox}{max totalsize={0.99\textwidth}{0.32\textheight},center}
    \usebox{\whenadapttreebox}
  \end{adjustbox}
  \vspace{-1.5em}
  \caption{Timing of Adaptation, organized by Input
    Visibility~$\times$~Update Persistence and orthogonal to
    Figure~\ref{fig:taxonomy}. Each leaf names representative
    methods; protocols spanning multiple regimes appear in more than one
    leaf. Adjacent no-update inference-time methods are shown for
  boundary clarity but are not counted as strict UPT.}
  \vspace{-0.5em}
  \label{fig:whenadapt}
\end{figure*}

The update-object taxonomy asks what internal object supplies the
signal. A timing view asks when target inputs become
visible and how long the induced change persists. We call these
two axes \emph{Input Visibility} and \emph{Update Persistence}.
They are orthogonal to family membership: the same update object
can be redeployed across regimes, so timing decides deployment
cost while family decides supervision shape. Cross-cutting these
two axes yields the regimes of Figure~\ref{fig:whenadapt}, which
charts adjacent no-update inference-time optimization that
uses similar internal signals but does not satisfy the
explicit-update requirement of strict UPT.

\vspace{-0.25em}
\paragraph{Pre-sample regimes.}
The first four regimes differ in how much of the target distribution
the update sees. \emph{Offline corpus UPT} is the broadest, covering
continued pretraining, instruction self-curation, reasoning
self-training, and offline
self-rewarding~\citep[e.g.,][]{ke2023cptlm,yuan2024selfrewarding,huang2026rzero};
the target distribution is invisible. \emph{Full-cohort transductive
adaptation} updates over the entire target cohort at
once~\citep{zuo2025ttrl,wei2025mmupt,wang2025scope,xing2025compass,zhao2026echo,wang2026selfharmony}.
\emph{Few-sample target adaptation} sees only a small slice and
generalizes to a held-out
remainder~\citep{singh2025ttrv,gao2025oneshot,prabhudesai2025rent,li2025rlsc};
this is a strictly stronger generalization claim than full-cohort.
\emph{Streaming continual adaptation} is properly online: at sample
$t$ the model uses only prefix $1{:}t{-}1$, with updates
accumulating
forward~\citep{hu2025test,singh2025ttrv,strich2026secl,liu2026ttvla}.

\vspace{-0.25em}
\paragraph{Within-sample regimes.}
The remaining two interleave adaptation with prediction.
\emph{Test-time instance adaptation} fits a small update for the
current instance and resets at its
boundary~\citep{jang2024tttnn,bansal2026qttt,hu2025slot,xu2025you,xu2026uldtta,kang2025modelwhisper}.
\emph{Within-sequence adaptation} goes finer still: the update
unfolds across chunks or token-states of one sequence and resets at
its boundary~\citep{feng2026inplace,sim2026ponderttt};
the update is to persistent local state, satisfying~\bcref{B1}.
Protocols can instantiate multiple regimes. Figure~\ref{fig:whenadapt}
records each realized protocol; Family~II\,$\times$\,full-cohort
transductive is the densest observed cell, exemplified by
\textsc{TTRL}.

\vspace{-0.25em}
\section{Challenges and Future Directions}
\vspace{-0.25em}
\label{sec:open-problems}

Recursive error propagation is the central open problem: updates
reinforce misordered outputs and shift the distribution from which the
next proxy is computed. The four families expose different links in
this chain and suggest distinct research priorities.

\vspace{-0.25em}
\paragraph{Separate signal quality from task structure.}
Consensus can be informative because errors cancel, but only after
an extractor defines which outputs agree. Boxed answers, finite
choice sets, and code signatures provide such structure without
determining correctness; execution and unit tests return a verdict
and therefore cross~\bcref{B3}. Future evaluations should report the
fields in Table~\ref{tab:task-prior-audit} and test whether gains
survive alternative canonicalizers and open-ended reformulations.
This matters most for Families~II--III; meanwhile,
open-ended methods such as \textsc{G-Zero}~\citep{huang2026gzeroselfplayopenendedgeneration}
probe the brittle boundary of equality-based consensus.

\vspace{-0.25em}
\paragraph{Interrupt confidence and majority amplification.}
Family~I can turn low entropy into overconfidence; Family~II can
turn a popular error into a training reward. Among the 22 Family~II
methods, 13 use hard majority signals, so evaluations should include
diversity and wrong-majority diagnostics. \textsc{EVOL-RL}~\citep{zhou2025evolrl}
adds semantic novelty, while \textsc{Dual~Consensus}~\citep{du2026dualconsensusescapingspurious}
and \textsc{CSRS}~\citep{yu2026stabilizingunsupervisedselfevolutionmllms}
soften or diversify consensus. \textsc{T$^3$RL}~\citep{liao2026t3rl}
shows that an independent execution channel can break false-popular
modes, motivating independent internal views for strict UPT.

\vspace{-0.25em}
\paragraph{Measure target and judge drift across rounds.}
Family~III freezes selected generations into training targets, so
confirmation bias can survive even when the next sampling round
looks more confident. Family~IV adds evaluator drift: bias
amplification, judge saturation, and weak chosen--rejected margins
are addressed separately by CREAM, Meta-Rewarding, and Temporal
SRLM~\citep{wang2024cream,wu2024metarewarding,jin2025temporalselfrewarding}.
Across these interventions, a common longitudinal protocol would
track target diversity, actor--judge correlation, held-out quality,
and perturbation recovery.

\vspace{-0.25em}
\paragraph{Control initialization and timing.}
Intrinsic-feedback results depend on the starting checkpoint and
update duration. \citet{zhang2025nofreelunch} show that these factors
materially change the outcome of intrinsic-feedback training.
Cross-family benchmarks should fix backbone, starting checkpoint,
target split, rollout and update budgets, task-structure prior, and
held-out evaluation to isolate the effects of signal family,
initialization, and timing.

\vspace{-0.25em}
\section{Conclusion}
\vspace{-0.25em}
\label{sec:conclusion}

\nopagebreak
We survey 80 strict UPT methods around a single question: which
model-derived object supplies the update signal? The resulting
taxonomy has four families, while Input Visibility and Update
Persistence provide an orthogonal map of when target inputs become
visible and how long updates persist. Together, these views clarify
UPT's central trade-off: more semantically expressive signals support
open-ended outputs but create longer feedback paths through which
proxy errors can be reinforced. Method selection and evaluation
should therefore align the update object with task structure and
deployment timing, and place safeguards where the proxy enters the
update.

\section*{Limitations}

The inventory is frozen in May~2026 and therefore excludes later
work. Representative results retain each source paper's backbone,
budget, metric, and adaptation protocol, supporting mechanism-level
synthesis rather than pooled effect-size estimation. Hybrid methods
follow the primary update object consumed by the gradient during
adaptation. Finally, \bcref{B1}--\bcref{B4} constrain the update rule rather than
the deployment stack. Because no external oracle scores the updated
model, reward hacking and silent per-problem degradation can
accumulate unobserved, so independent held-out evaluation,
monitoring, and red-teaming remain necessary and are not excluded by
the strict boundary; we organize such safeguards by feedback path
(\S\ref{sec:synthesis}) but do not evaluate their effectiveness,
which requires controlled study.

\section*{Acknowledgments}
This work was supported in part by the National Natural Science Foundation of China (Grant Nos. 92370204 and 62506318); in part by the National Key R\&D Program of China (Grant No. 2023YFF0725001); in part by the Guangdong Provincial Key Laboratory of Frontier Basic Science for All-domain Intelligence; in part by the Guangdong Basic and Applied Basic Research Foundation (Grant No. 2023B1515120057); in part by the Key-Area Special Project of Guangdong Provincial Ordinary Universities (Grant No. 2024ZDZX1007); in part by the Guangdong Provincial Department of Education Project (Grant No. 2024KQNCX028); in part by the Scientific Research Projects for the Higher-educational Institutions, Education Bureau of Guangzhou Municipality (Grant No. 2024312096); and in part by the Guangzhou-HKUST(GZ) Joint Funding Program, Education Bureau of Guangzhou Municipality (Grant No. 2025A03J3957).

\bibliography{references_new}

@inproceedings{abbes2025replayalign,
  title     = {Revisiting Replay and Gradient Alignment For Continual Pretraining of Large Language Models},
  author    = {Istabrak Abbes and Gopeshh Subbaraj and Matthew Riemer and Nizar Islah and Tsuguchika Tabaru and Hiroaki Kingetsu and Sarath Chandar and Irina Rish},
  booktitle = {Proceedings of The 4th Conference on Lifelong Learning Agents},
  pages     = {465--486},
  year      = {2026},
  volume    = {330},
  series    = {Proceedings of Machine Learning Research},
  publisher = {PMLR},
  url       = {https://proceedings.mlr.press/v330/abbes26a.html}
}

@inproceedings{agarwal2025unreasonable,
  title     = {The Unreasonable Effectiveness of Entropy Minimization in {LLM} Reasoning},
  author    = {Shivam Agarwal and Zimin Zhang and Lifan Yuan and Jiawei Han and Hao Peng},
  booktitle = {The Thirty-ninth Annual Conference on Neural Information Processing Systems},
  year      = {2025},
  url       = {https://openreview.net/forum?id=UfFTBEsLgI}
}

@inproceedings{arazo2020pseudolabeling,
  title     = {Pseudo-Labeling and Confirmation Bias in Deep Semi-Supervised Learning},
  author    = {Eric Arazo and Diego Ortego and Paul Albert and Noel E. O'Connor and Kevin McGuinness},
  booktitle = {2020 International Joint Conference on Neural Networks},
  pages     = {1--8},
  year      = {2020},
  doi       = {10.1109/IJCNN48605.2020.9207304},
  url       = {https://arxiv.org/abs/1908.02983}
}

@inproceedings{wei2021theoreticalselftraining,
  title     = {Theoretical Analysis of Self-Training with Deep Networks on Unlabeled Data},
  author    = {Colin Wei and Kendrick Shen and Yining Chen and Tengyu Ma},
  booktitle = {International Conference on Learning Representations},
  year      = {2021},
  url       = {https://openreview.net/forum?id=rC8sJ4i6kaH}
}

@inproceedings{arzhantsev2025roirl,
  title     = {Roi{RL}: Efficient, Self-Supervised Reasoning with Offline Iterative Reinforcement Learning},
  author    = {Aleksei Arzhantsev and Otmane Sakhi and Flavian Vasile},
  booktitle = {NeurIPS 2025 Workshop on Efficient Reasoning},
  year      = {2025},
  url       = {https://openreview.net/forum?id=PeJ1eGGygZ}
}

@misc{bansal2026qttt,
  title         = {Let's (not) just put things in Context: Test-Time Training for Long-Context LLMs},
  author        = {Rachit Bansal and Aston Zhang and Rishabh Tiwari and Lovish Madaan and Sai Surya Duvvuri and Devvrit Khatri and David Brandfonbrener and David Alvarez-Melis and Prajjwal Bhargava and Mihir Sanjay Kale and Samy Jelassi},
  year          = {2025},
  eprint        = {2512.13898},
  archiveprefix = {arXiv},
  primaryclass  = {cs.LG},
  url           = {https://arxiv.org/abs/2512.13898}
}

@misc{chen2025longcotsurvey,
  title         = {Towards Reasoning Era: A Survey of Long Chain-of-Thought for Reasoning Large Language Models},
  author        = {Qiguang Chen and Libo Qin and Jinhao Liu and Dengyun Peng and Jiannan Guan and Peng Wang and Mengkang Hu and Yuhang Zhou and Te Gao and Wanxiang Che},
  year          = {2025},
  eprint        = {2503.09567},
  archiveprefix = {arXiv},
  primaryclass  = {cs.AI},
  url           = {https://arxiv.org/abs/2503.09567}
}

@inproceedings{chen2025longpo,
  title     = {Long{PO}: Long Context Self-Evolution of Large Language Models through Short-to-Long Preference Optimization},
  author    = {Guanzheng Chen and Xin Li and Michael Shieh and Lidong Bing},
  booktitle = {The Thirteenth International Conference on Learning Representations},
  year      = {2025},
  url       = {https://openreview.net/forum?id=qTrEq31Shm}
}

@misc{du2026dare,
  title         = {Distribution-Aware Reward Estimation for Test-Time Reinforcement Learning},
  author        = {Bodong Du and Xuanqi Huang and Xiaomeng Li},
  year          = {2026},
  eprint        = {2601.21804},
  archiveprefix = {arXiv},
  primaryclass  = {cs.CL},
  url           = {https://arxiv.org/abs/2601.21804}
}

@misc{du2026dualconsensusescapingspurious,
  title         = {Dual Consensus: Escaping from Spurious Majority in Unsupervised RLVR via Two-Stage Vote Mechanism},
  author        = {Kaixuan Du and Meng Cao and Hang Zhang and Yukun Wang and Xiangzhou Huang and Ni Li},
  year          = {2026},
  eprint        = {2603.16223},
  archiveprefix = {arXiv},
  primaryclass  = {cs.LG},
  url           = {https://arxiv.org/abs/2603.16223}
}

@inproceedings{elhady2025languageadapt,
  title     = {Emergent Abilities of Large Language Models under Continued Pre-training for Language Adaptation},
  author    = {Elhady, Ahmed  and
               Agirre, Eneko  and
               Artetxe, Mikel},
  editor    = {Che, Wanxiang  and
               Nabende, Joyce  and
               Shutova, Ekaterina  and
               Pilehvar, Mohammad Taher},
  booktitle = {Proceedings of the 63rd Annual Meeting of the Association for Computational Linguistics (Volume 1: Long Papers)},
  month     = jul,
  year      = {2025},
  address   = {Vienna, Austria},
  publisher = {Association for Computational Linguistics},
  url       = {https://aclanthology.org/2025.acl-long.1547/},
  doi       = {10.18653/v1/2025.acl-long.1547},
  pages     = {32174--32186},
  isbn      = {979-8-89176-251-0}
}

@inproceedings{feng2026inplace,
  title     = {In-Place Test-Time Training},
  author    = {Guhao Feng and Shengjie Luo and Kai Hua and Ge Zhang and Wenhao Huang and Di He and Tianle Cai},
  booktitle = {The Fourteenth International Conference on Learning Representations},
  year      = {2026},
  url       = {https://openreview.net/forum?id=dTWfCLSoyl}
}

@misc{fu2024dataengineering,
  title         = {Data Engineering for Scaling Language Models to 128K Context},
  author        = {Yao Fu and Rameswar Panda and Xinyao Niu and Xiang Yue and Hannaneh Hajishirzi and Yoon Kim and Hao Peng},
  year          = {2024},
  eprint        = {2402.10171},
  archiveprefix = {arXiv},
  primaryclass  = {cs.CL},
  url           = {https://arxiv.org/abs/2402.10171}
}

@misc{gao2025longmagpie,
  title         = {LongMagpie: A Self-synthesis Method for Generating Large-scale Long-context Instructions},
  author        = {Chaochen Gao and Xing Wu and Zijia Lin and Debing Zhang and Songlin Hu},
  year          = {2025},
  eprint        = {2505.17134},
  archiveprefix = {arXiv},
  primaryclass  = {cs.CL},
  url           = {https://arxiv.org/abs/2505.17134}
}

@misc{gao2025oneshot,
  title         = {One-shot Entropy Minimization},
  author        = {Zitian Gao and Lynx Chen and Haoming Luo and Joey Zhou and Bryan Dai},
  year          = {2025},
  eprint        = {2505.20282},
  archiveprefix = {arXiv},
  primaryclass  = {cs.CL},
  url           = {https://arxiv.org/abs/2505.20282},
  note          = {Work in progress}
}

@misc{gao2026aeroautonomousevolutionaryreasoning,
  title         = {AERO: Autonomous Evolutionary Reasoning Optimization via Endogenous Dual-Loop Feedback},
  author        = {Zhitao Gao and Jie Ma and Xuhong Li and Pengyu Li and Ning Qu and Yaqiang Wu and Hui Liu and Jun Liu},
  year          = {2026},
  eprint        = {2602.03084},
  archiveprefix = {arXiv},
  primaryclass  = {cs.CL},
  url           = {https://arxiv.org/abs/2602.03084}
}

@inproceedings{guo2025stabilitygap,
  title     = {Efficient Domain Continual pretraining by Mitigating the Stability Gap},
  author    = {Guo, Yiduo  and
               Fu, Jie  and
               Zhang, Huishuai  and
               Zhao, Dongyan},
  editor    = {Che, Wanxiang  and
               Nabende, Joyce  and
               Shutova, Ekaterina  and
               Pilehvar, Mohammad Taher},
  booktitle = {Proceedings of the 63rd Annual Meeting of the Association for Computational Linguistics (Volume 1: Long Papers)},
  month     = jul,
  year      = {2025},
  address   = {Vienna, Austria},
  publisher = {Association for Computational Linguistics},
  url       = {https://aclanthology.org/2025.acl-long.1578/},
  doi       = {10.18653/v1/2025.acl-long.1578},
  pages     = {32850--32870},
  isbn      = {979-8-89176-251-0}
}

@misc{he2026ttsr,
  title         = {TTSR: Test-Time Self-Reflection for Continual Reasoning Improvement},
  author        = {Haoyang He and Zihua Rong and Liangjie Zhao and Yunjia Zhao and Lan Yang and Honggang Zhang},
  year          = {2026},
  eprint        = {2603.03297},
  archiveprefix = {arXiv},
  primaryclass  = {cs.CL},
  url           = {https://arxiv.org/abs/2603.03297}
}

@misc{hong2025suderselfimprovingunifiedlarge,
  title         = {SUDER: Self-Improving Unified Large Multimodal Models for Understanding and Generation with Dual Self-Rewards},
  author        = {Jixiang Hong and Yiran Zhang and Guanzhong Wang and Yi Liu and Ji-Rong Wen and Rui Yan},
  year          = {2025},
  eprint        = {2506.07963},
  archiveprefix = {arXiv},
  primaryclass  = {cs.AI},
  url           = {https://arxiv.org/abs/2506.07963}
}

@misc{hu2025slot,
  title         = {SLOT: Sample-specific Language Model Optimization at Test-time},
  author        = {Yang Hu and Xingyu Zhang and Xueji Fang and Zhiyang Chen and Xiao Wang and Huatian Zhang and Guojun Qi},
  year          = {2025},
  eprint        = {2505.12392},
  archiveprefix = {arXiv},
  primaryclass  = {cs.CL},
  url           = {https://arxiv.org/abs/2505.12392}
}

@inproceedings{hu2025test,
  title     = {Test-Time Learning for Large Language Models},
  author    = {Hu, Jinwu and Zhang, Zitian and Chen, Guohao and Wen, Xutao and Shuai, Chao and Luo, Wei and Xiao, Bin and Li, Yuanqing and Tan, Mingkui},
  booktitle = {Proceedings of the 42nd International Conference on Machine Learning},
  pages     = {24823--24849},
  year      = {2025},
  editor    = {Singh, Aarti and Fazel, Maryam and Hsu, Daniel and Lacoste-Julien, Simon and Berkenkamp, Felix and Maharaj, Tegan and Wagstaff, Kiri and Zhu, Jerry},
  volume    = {267},
  series    = {Proceedings of Machine Learning Research},
  month     = {13--19 Jul},
  publisher = {PMLR},
  url       = {https://proceedings.mlr.press/v267/hu25z.html}
}

@inproceedings{huang2023selfimprove,
  title     = {Large Language Models Can Self-Improve},
  author    = {Huang, Jiaxin  and
               Gu, Shixiang  and
               Hou, Le  and
               Wu, Yuexin  and
               Wang, Xuezhi  and
               Yu, Hongkun  and
               Han, Jiawei},
  editor    = {Bouamor, Houda  and
               Pino, Juan  and
               Bali, Kalika},
  booktitle = {Proceedings of the 2023 Conference on Empirical Methods in Natural Language Processing},
  month     = dec,
  year      = {2023},
  address   = {Singapore},
  publisher = {Association for Computational Linguistics},
  url       = {https://aclanthology.org/2023.emnlp-main.67/},
  doi       = {10.18653/v1/2023.emnlp-main.67},
  pages     = {1051--1068}
}

@misc{huang2026gzeroselfplayopenendedgeneration,
  title         = {G-Zero: Self-Play for Open-Ended Generation from Zero Data},
  author        = {Chengsong Huang and Haolin Liu and Tong Zheng and Runpeng Dai and Langlin Huang and Jinyuan Li and Zongxia Li and Zhepei Wei and Yu Meng and Jiaxin Huang},
  year          = {2026},
  eprint        = {2605.09959},
  archiveprefix = {arXiv},
  primaryclass  = {cs.LG},
  url           = {https://arxiv.org/abs/2605.09959}
}

@misc{huang2026rzero,
  title         = {R-Zero: Self-Evolving Reasoning LLM from Zero Data},
  author        = {Chengsong Huang and Wenhao Yu and Xiaoyang Wang and Hongming Zhang and Zongxia Li and Ruosen Li and Jiaxin Huang and Haitao Mi and Dong Yu},
  year          = {2025},
  eprint        = {2508.05004},
  archiveprefix = {arXiv},
  primaryclass  = {cs.LG},
  url           = {https://arxiv.org/abs/2508.05004}
}

@inproceedings{jang2024tttnn,
  title     = {Test-Time Training on Nearest Neighbors for Large Language Models},
  author    = {Moritz Hardt and Yu Sun},
  booktitle = {The Twelfth International Conference on Learning Representations},
  year      = {2024},
  url       = {https://openreview.net/forum?id=CNL2bku4ra}
}

@misc{jin2025temporalselfrewarding,
  title         = {Temporal Self-Rewarding Language Models: Decoupling Chosen-Rejected via Past-Future},
  author        = {Yidong Wang and Xin Wang and Cunxiang Wang and Junfeng Fang and Qiufeng Wang and Jianing Chu and Xuran Meng and Shuxun Yang and Libo Qin and Yue Zhang and Wei Ye and Shikun Zhang},
  year          = {2025},
  eprint        = {2508.06026},
  archiveprefix = {arXiv},
  primaryclass  = {cs.CL},
  url           = {https://arxiv.org/abs/2508.06026}
}

@misc{kang2025modelwhisper,
  title         = {Model Whisper: Steering Vectors Unlock Large Language Models' Potential in Test-time},
  author        = {Xinyue Kang and Diwei Shi and Li Chen},
  year          = {2025},
  eprint        = {2512.04748},
  archiveprefix = {arXiv},
  primaryclass  = {cs.CL},
  url           = {https://arxiv.org/abs/2512.04748}
}

@misc{ke2023cptlm,
  title         = {Continual Pre-training of Language Models},
  author        = {Zixuan Ke and Yijia Shao and Haowei Lin and Tatsuya Konishi and Gyuhak Kim and Bing Liu},
  year          = {2023},
  eprint        = {2302.03241},
  archiveprefix = {arXiv},
  primaryclass  = {cs.CL},
  url           = {https://arxiv.org/abs/2302.03241}
}

@misc{kumar2025posttraining,
  title         = {LLM Post-Training: A Deep Dive into Reasoning Large Language Models},
  author        = {Komal Kumar and Tajamul Ashraf and Omkar Thawakar and Rao Muhammad Anwer and Hisham Cholakkal and Mubarak Shah and Ming-Hsuan Yang and Phillip H.S. Torr and Fahad Shahbaz Khan and Salman Khan},
  year          = {2025},
  eprint        = {2502.21321},
  archiveprefix = {arXiv},
  primaryclass  = {cs.CL},
  url           = {https://arxiv.org/abs/2502.21321}
}

@misc{li2024instructionbacktranslation,
  title         = {Self-Alignment with Instruction Backtranslation},
  author        = {Xian Li and Ping Yu and Chunting Zhou and Timo Schick and Omer Levy and Luke Zettlemoyer and Jason Weston and Mike Lewis},
  year          = {2023},
  eprint        = {2308.06259},
  archiveprefix = {arXiv},
  primaryclass  = {cs.CL},
  url           = {https://arxiv.org/abs/2308.06259}
}

@misc{li2025rlsc,
  title         = {Confidence Is All You Need: Few-Shot RL Fine-Tuning of Language Models},
  author        = {Pengyi Li and Matvey Skripkin and Alexander Zubrey and Andrey Kuznetsov and Ivan Oseledets},
  year          = {2025},
  eprint        = {2506.06395},
  archiveprefix = {arXiv},
  primaryclass  = {cs.CL},
  url           = {https://arxiv.org/abs/2506.06395}
}

@misc{liang2025ttasurvey,
  title         = {A Comprehensive Survey on Test-Time Adaptation under Distribution Shifts},
  author        = {Jian Liang and Ran He and Tieniu Tan},
  year          = {2023},
  eprint        = {2303.15361},
  archiveprefix = {arXiv},
  primaryclass  = {cs.LG},
  url           = {https://arxiv.org/abs/2303.15361}
}

@misc{liao2026t3rl,
  title         = {Tool Verification for Test-Time Reinforcement Learning},
  author        = {Ruotong Liao and Nikolai Röhrich and Xiaohan Wang and Yuhui Zhang and Yasaman Samadzadeh and Volker Tresp and Serena Yeung-Levy},
  year          = {2026},
  eprint        = {2603.02203},
  archiveprefix = {arXiv},
  primaryclass  = {cs.AI},
  url           = {https://arxiv.org/abs/2603.02203}
}

@inproceedings{liu2024e2llm,
  title     = {E2-{LLM}: Efficient and Extreme Length Extension of Large Language Models},
  author    = {Liu, Jiaheng  and
               Bai, Zhiqi  and
               Zhang, Yuanxing  and
               Zhang, Chenchen  and
               Zhang, Yu  and
               Zhang, Ge  and
               Wang, Jiakai  and
               Que, Haoran  and
               Chen, Yukang  and
               Su, Wenbo  and
               Ge, Tiezheng  and
               Fu, Jie  and
               Chen, Wenhu  and
               Zheng, Bo},
  editor    = {Ku, Lun-Wei  and
               Martins, Andre  and
               Srikumar, Vivek},
  booktitle = {Findings of the Association for Computational Linguistics: ACL 2024},
  month     = aug,
  year      = {2024},
  address   = {Bangkok, Thailand},
  publisher = {Association for Computational Linguistics},
  url       = {https://aclanthology.org/2024.findings-acl.252/},
  doi       = {10.18653/v1/2024.findings-acl.252},
  pages     = {4243--4253}
}

@inproceedings{liu2025dte,
  title     = {{DEBATE}, {TRAIN}, {EVOLVE}: {S}elf{-}{E}volution of Language Model Reasoning},
  author    = {Srivastava, Gaurav  and
               Bi, Zhenyu  and
               Lu, Meng  and
               Wang, Xuan},
  editor    = {Christodoulopoulos, Christos  and
               Chakraborty, Tanmoy  and
               Rose, Carolyn  and
               Peng, Violet},
  booktitle = {Proceedings of the 2025 Conference on Empirical Methods in Natural Language Processing},
  month     = nov,
  year      = {2025},
  address   = {Suzhou, China},
  publisher = {Association for Computational Linguistics},
  url       = {https://aclanthology.org/2025.emnlp-main.1666/},
  doi       = {10.18653/v1/2025.emnlp-main.1666},
  pages     = {32764--32810},
  isbn      = {979-8-89176-332-6}
}

@misc{liu2025ettrl,
  title         = {ETTRL: Balancing Exploration and Exploitation in LLM Test-Time Reinforcement Learning Via Entropy Mechanism},
  author        = {Jia Liu and ChangYi He and YingQiao Lin and MingMin Yang and FeiYang Shen and ShaoGuo Liu},
  year          = {2025},
  eprint        = {2508.11356},
  archiveprefix = {arXiv},
  primaryclass  = {cs.LG},
  url           = {https://arxiv.org/abs/2508.11356}
}

@misc{liu2026ttvla,
  title         = {On-the-Fly VLA Adaptation via Test-Time Reinforcement Learning},
  author        = {Changyu Liu and Yiyang Liu and Taowen Wang and Qiao Zhuang and James Chenhao Liang and Wenhao Yang and Renjing Xu and Qifan Wang and Dongfang Liu and Cheng Han},
  year          = {2026},
  eprint        = {2601.06748},
  archiveprefix = {arXiv},
  primaryclass  = {cs.RO},
  url           = {https://arxiv.org/abs/2601.06748}
}

@misc{moradi2026disctt,
  title         = {DiSCTT: Consensus-Guided Self-Curriculum for Efficient Test-Time Adaptation in Reasoning},
  author        = {Mohammad Mahdi Moradi and Sudhir Mudur},
  year          = {2026},
  eprint        = {2603.05357},
  archiveprefix = {arXiv},
  primaryclass  = {cs.CL},
  url           = {https://arxiv.org/abs/2603.05357}
}

@inproceedings{pan2026learninggenerateunderstandingunderstandingdriven,
  author    = {Pan, Jiadong and Li, Liang and Peng, Yuxin and Tang, Yu-Ming and Wang, Shuohuan and Sun, Yu and Wu, Hua and Huang, Qingming and Wang, Haifeng},
  title     = {Learning to Generate via Understanding: Understanding-Driven Intrinsic Rewarding for Unified Multimodal Models},
  booktitle = {Proceedings of the IEEE/CVF Conference on Computer Vision and Pattern Recognition},
  month     = {June},
  year      = {2026},
  pages     = {22174-22184}
}

@misc{prabhudesai2025rent,
  title         = {Maximizing Confidence Alone Improves Reasoning},
  author        = {Mihir Prabhudesai and Lili Chen and Alex Ippoliti and Katerina Fragkiadaki and Hao Liu and Deepak Pathak},
  year          = {2025},
  eprint        = {2505.22660},
  archiveprefix = {arXiv},
  primaryclass  = {cs.LG},
  url           = {https://arxiv.org/abs/2505.22660}
}

@inproceedings{prasad2025scpo,
  title     = {Self-Consistency Preference Optimization},
  author    = {Prasad, Archiki and Yuan, Weizhe and Pang, Richard Yuanzhe and Xu, Jing and Fazel-Zarandi, Maryam and Bansal, Mohit and Sukhbaatar, Sainbayar and Weston, Jason E and Yu, Jane},
  booktitle = {Proceedings of the 42nd International Conference on Machine Learning},
  pages     = {49737--49751},
  year      = {2025},
  editor    = {Singh, Aarti and Fazel, Maryam and Hsu, Daniel and Lacoste-Julien, Simon and Berkenkamp, Felix and Maharaj, Tegan and Wagstaff, Kiri and Zhu, Jerry},
  volume    = {267},
  series    = {Proceedings of Machine Learning Research},
  month     = {13--19 Jul},
  publisher = {PMLR},
  url       = {https://proceedings.mlr.press/v267/prasad25a.html}
}

@misc{qian2024simplescalable,
  title         = {Simple and Scalable Strategies to Continually Pre-train Large Language Models},
  author        = {Adam Ibrahim and Benjamin Thérien and Kshitij Gupta and Mats L. Richter and Quentin Anthony and Timothée Lesort and Eugene Belilovsky and Irina Rish},
  year          = {2024},
  eprint        = {2403.08763},
  archiveprefix = {arXiv},
  primaryclass  = {cs.LG},
  url           = {https://arxiv.org/abs/2403.08763}
}

@misc{rentschler2026rlme,
  title         = {Reinforcement Learning from Meta-Evaluation: Aligning Language Models Without Ground-Truth Labels},
  author        = {Micah Rentschler and Jesse Roberts},
  year          = {2026},
  eprint        = {2601.21268},
  archiveprefix = {arXiv},
  primaryclass  = {cs.NE},
  url           = {https://arxiv.org/abs/2601.21268}
}

@misc{samanta2025maca,
  title         = {Self-Improvement of Language Models by Post-Training on Multi-Agent Debate},
  author        = {Ankur Samanta and Akshayaa Magesh and Runzhe Wu and Ayush Jain and Youliang Yu and Daniel Jiang and Boris Vidolov and Paul Sajda and Yonathan Efroni and Kaveh Hassani},
  year          = {2025},
  eprint        = {2509.15172},
  archiveprefix = {arXiv},
  primaryclass  = {cs.AI},
  url           = {https://arxiv.org/abs/2509.15172}
}

@misc{shao2024deepseekmath,
  title         = {DeepSeekMath: Pushing the Limits of Mathematical Reasoning in Open Language Models},
  author        = {Zhihong Shao and Peiyi Wang and Qihao Zhu and Runxin Xu and Junxiao Song and Xiao Bi and Haowei Zhang and Mingchuan Zhang and Y.K. Li and Y. Wu and Daya Guo},
  year          = {2024},
  eprint        = {2402.03300},
  archiveprefix = {arXiv},
  primaryclass  = {cs.CL},
  url           = {https://arxiv.org/abs/2402.03300}
}

@inproceedings{shen2025cycleinstruct,
  title     = {{CYCLE}-{INSTRUCT}: Fully Seed-Free Instruction Tuning via Dual Self-Training and Cycle Consistency},
  author    = {Shen, Zhanming  and
               Chen, Hao  and
               Tang, Yulei  and
               Zhu, Shaolin  and
               Ye, Wentao  and
               Hu, Xiaomeng  and
               Wang, Haobo  and
               Chen, Gang  and
               Zhao, Junbo},
  editor    = {Christodoulopoulos, Christos  and
               Chakraborty, Tanmoy  and
               Rose, Carolyn  and
               Peng, Violet},
  booktitle = {Proceedings of the 2025 Conference on Empirical Methods in Natural Language Processing},
  month     = nov,
  year      = {2025},
  address   = {Suzhou, China},
  publisher = {Association for Computational Linguistics},
  url       = {https://aclanthology.org/2025.emnlp-main.258/},
  doi       = {10.18653/v1/2025.emnlp-main.258},
  pages     = {5123--5137},
  isbn      = {979-8-89176-332-6}
}

@misc{shi2025lrmselftrain,
  title         = {Can Large Reasoning Models Self-Train?},
  author        = {Sheikh Shafayat and Fahim Tajwar and Ruslan Salakhutdinov and Jeff Schneider and Andrea Zanette},
  year          = {2025},
  eprint        = {2505.21444},
  archiveprefix = {arXiv},
  primaryclass  = {cs.LG},
  url           = {https://arxiv.org/abs/2505.21444}
}

@misc{sim2026ponderttt,
  title         = {When to Ponder: Adaptive Compute Allocation for Code Generation via Test-Time Training},
  author        = {Gihyeon Sim},
  year          = {2025},
  eprint        = {2601.00894},
  archiveprefix = {arXiv},
  primaryclass  = {cs.LG},
  url           = {https://arxiv.org/abs/2601.00894}
}

@misc{singh2025ttrv,
  title         = {TTRV: Test-Time Reinforcement Learning for Vision Language Models},
  author        = {Akshit Singh and Shyam Marjit and Wei Lin and Paul Gavrikov and Serena Yeung-Levy and Hilde Kuehne and Rogerio Feris and Sivan Doveh and James Glass and M. Jehanzeb Mirza},
  year          = {2025},
  eprint        = {2510.06783},
  archiveprefix = {arXiv},
  primaryclass  = {cs.CV},
  url           = {https://arxiv.org/abs/2510.06783}
}

@misc{song2026queryconditionedtesttimeselftraininglarge,
  title         = {Query-Conditioned Test-Time Self-Training for Large Language Models},
  author        = {Chaehee Song and Minseok Seo and Yeeun Seong and Doyi Kim and Changick Kim},
  year          = {2026},
  eprint        = {2605.13369},
  archiveprefix = {arXiv},
  primaryclass  = {cs.CL},
  url           = {https://arxiv.org/abs/2605.13369}
}

@misc{strich2026secl,
  title         = {Self-Calibrating Language Models via Test-Time Discriminative Distillation},
  author        = {Mohamed Rissal Hedna and Jan Strich and Martin Semmann and Chris Biemann},
  year          = {2026},
  eprint        = {2604.09624},
  archiveprefix = {arXiv},
  primaryclass  = {cs.CL},
  url           = {https://arxiv.org/abs/2604.09624}
}

@inproceedings{sui2026conl,
  title         = {Conversation for Non-verifiable Learning: Self-Evolving {LLM}s through Meta-Evaluation},
  author        = {Yuan Sui and Bryan Hooi},
  booktitle     = {Proceedings of the 43rd International Conference on Machine Learning},
  year          = {2026},
  volume        = {306},
  series        = {Proceedings of Machine Learning Research},
  eprint        = {2601.21464},
  archiveprefix = {arXiv},
  url           = {https://arxiv.org/abs/2601.21464}
}

@misc{tan2026metattrl,
  title         = {Meta-TTRL: A Metacognitive Framework for Self-Improving Test-Time Reinforcement Learning in Unified Multimodal Models},
  author        = {Lit Sin Tan and Junzhe Chen and Xiaolong Fu and Lichen Ma and Junshi Huang and Jianzhong Shi and Yan Li and Lijie Wen},
  year          = {2026},
  eprint        = {2603.15724},
  archiveprefix = {arXiv},
  primaryclass  = {cs.LG},
  url           = {https://arxiv.org/abs/2603.15724}
}

@misc{tao2024selfevolution,
  title         = {A Survey on Self-Evolution of Large Language Models},
  author        = {Zhengwei Tao and Ting-En Lin and Xiancai Chen and Hangyu Li and Yuchuan Wu and Yongbin Li and Zhi Jin and Fei Huang and Dacheng Tao and Jingren Zhou},
  year          = {2024},
  eprint        = {2404.14387},
  archiveprefix = {arXiv},
  primaryclass  = {cs.CL},
  url           = {https://arxiv.org/abs/2404.14387}
}

@misc{thawakar2025evolmm,
  title         = {EvoLMM: Self-Evolving Large Multimodal Models with Continuous Rewards},
  author        = {Omkar Thawakar and Shravan Venkatraman and Ritesh Thawkar and Abdelrahman Shaker and Hisham Cholakkal and Rao Muhammad Anwer and Salman Khan and Fahad Khan},
  year          = {2025},
  eprint        = {2511.16672},
  archiveprefix = {arXiv},
  primaryclass  = {cs.CV},
  url           = {https://arxiv.org/abs/2511.16672}
}

@misc{vanniekerk2025rlsf,
  title         = {Post-Training Large Language Models via Reinforcement Learning from Self-Feedback},
  author        = {Carel van Niekerk and Renato Vukovic and Benjamin Matthias Ruppik and Hsien-chin Lin and Milica Ga{\v{s}}i{\'c}},
  year          = {2025},
  eprint        = {2507.21931},
  archiveprefix = {arXiv},
  primaryclass  = {cs.CL},
  url           = {https://arxiv.org/abs/2507.21931}
}

@inproceedings{wang2023selfinstruct,
  title     = {Self-Instruct: Aligning Language Models with Self-Generated Instructions},
  author    = {Wang, Yizhong  and
               Kordi, Yeganeh  and
               Mishra, Swaroop  and
               Liu, Alisa  and
               Smith, Noah A.  and
               Khashabi, Daniel  and
               Hajishirzi, Hannaneh},
  editor    = {Rogers, Anna  and
               Boyd-Graber, Jordan  and
               Okazaki, Naoaki},
  booktitle = {Proceedings of the 61st Annual Meeting of the Association for Computational Linguistics (Volume 1: Long Papers)},
  month     = jul,
  year      = {2023},
  address   = {Toronto, Canada},
  publisher = {Association for Computational Linguistics},
  url       = {https://aclanthology.org/2023.acl-long.754/},
  doi       = {10.18653/v1/2023.acl-long.754},
  pages     = {13484--13508}
}

@inproceedings{wang2024cream,
  title     = {{CREAM}: Consistency Regularized Self-Rewarding Language Models},
  author    = {Zhaoyang Wang and Weilei He and Zhiyuan Liang and Xuchao Zhang and Chetan Bansal and Ying Wei and Weitong Zhang and Huaxiu Yao},
  booktitle = {The Thirteenth International Conference on Learning Representations},
  year      = {2025},
  url       = {https://openreview.net/forum?id=Vf6RDObyEF}
}

@misc{wang2024longselfimprove,
  title         = {Large Language Models Can Self-Improve in Long-context Reasoning},
  author        = {Siheng Li and Cheng Yang and Zesen Cheng and Lemao Liu and Mo Yu and Yujiu Yang and Wai Lam},
  year          = {2024},
  eprint        = {2411.08147},
  archiveprefix = {arXiv},
  primaryclass  = {cs.CL},
  url           = {https://arxiv.org/abs/2411.08147}
}

@inproceedings{wang2025concisereasoning,
  title     = {Self-Training Elicits Concise Reasoning in Large Language Models},
  author    = {Munkhbat, Tergel  and
               Ho, Namgyu  and
               Kim, Seo Hyun  and
               Yang, Yongjin  and
               Kim, Yujin  and
               Yun, Se-Young},
  editor    = {Che, Wanxiang  and
               Nabende, Joyce  and
               Shutova, Ekaterina  and
               Pilehvar, Mohammad Taher},
  booktitle = {Findings of the Association for Computational Linguistics: ACL 2025},
  month     = jul,
  year      = {2025},
  address   = {Vienna, Austria},
  publisher = {Association for Computational Linguistics},
  url       = {https://aclanthology.org/2025.findings-acl.1289/},
  doi       = {10.18653/v1/2025.findings-acl.1289},
  pages     = {25127--25152},
  isbn      = {979-8-89176-256-5}
}

@inproceedings{wang2025confidentreasoning,
  title     = {Self-Training Large Language Models with Confident Reasoning},
  author    = {Jang, Hyosoon  and
               Jang, Yunhui  and
               Lee, Sungjae  and
               Ok, Jungseul  and
               Ahn, Sungsoo},
  editor    = {Christodoulopoulos, Christos  and
               Chakraborty, Tanmoy  and
               Rose, Carolyn  and
               Peng, Violet},
  booktitle = {Findings of the Association for Computational Linguistics: EMNLP 2025},
  month     = nov,
  year      = {2025},
  address   = {Suzhou, China},
  publisher = {Association for Computational Linguistics},
  url       = {https://aclanthology.org/2025.findings-emnlp.806/},
  doi       = {10.18653/v1/2025.findings-emnlp.806},
  pages     = {14925--14939},
  isbn      = {979-8-89176-335-7}
}

@misc{wang2025scope,
  title         = {Beyond Majority Voting: Towards Fine-grained and More Reliable Reward Signal for Test-Time Reinforcement Learning},
  author        = {Weiqin Wang and Yile Wang and Kehao Chen and Hui Huang},
  year          = {2025},
  eprint        = {2512.15146},
  archiveprefix = {arXiv},
  primaryclass  = {cs.CL},
  url           = {https://arxiv.org/abs/2512.15146}
}

@inproceedings{wang2026selfharmony,
  title     = {{SELF}-{HARMONY}: {LEARNING} {TO} {HARMONIZE} {SELF}-{SUPERVISION} {AND} {SELF}-{PLAY} {IN} {TEST}-{TIME} {REINFORCEMENT} {LEARNING}},
  author    = {Ru Wang and Wei Huang and Qi Cao and Yusuke Iwasawa and Yutaka Matsuo and Jiaxian Guo},
  booktitle = {The Fourteenth International Conference on Learning Representations},
  year      = {2026},
  url       = {https://openreview.net/forum?id=ZzG6oJ5ehI}
}

@misc{wang2026vzeroselfimprovingmultimodalreasoning,
  title         = {V-Zero: Self-Improving Multimodal Reasoning with Zero Annotation},
  author        = {Han Wang and Yi Yang and Jingyuan Hu and Minfeng Zhu and Wei Chen},
  year          = {2026},
  eprint        = {2601.10094},
  archiveprefix = {arXiv},
  primaryclass  = {cs.CV},
  url           = {https://arxiv.org/abs/2601.10094}
}

@inproceedings{wei2025mmupt,
  title     = {First {SFT}, Second {RL}, Third {UPT}: Continual Improving Multi-Modal {LLM} Reasoning via Unsupervised Post-Training},
  author    = {Lai Wei and Yuting Li and Chen Wang and Yue Wang and Linghe Kong and Weiran Huang and Lichao Sun},
  booktitle = {Advances in Neural Information Processing Systems},
  volume    = {38},
  year      = {2025},
  doi       = {10.52202/085713-2084},
  url       = {https://proceedings.neurips.cc/paper_files/paper/2025/hash/59ddfff7979b43f54690fa986c0e5138-Abstract-Conference.html}
}

@inproceedings{wen2026selfevolvingvisionlanguagemodelsimage,
  title     = {Self-Evolving Vision-Language Models for Image Quality Assessment via Voting and Ranking},
  author    = {Wen Wen and tianwu zhi and Kanglong FAN and Yang Li and Xinge Peng and Yabin ZHANG and Yiting Liao and Junlin Li and Li zhang},
  booktitle = {The Fourteenth International Conference on Learning Representations},
  year      = {2026},
  url       = {https://openreview.net/forum?id=INOi0YqI8p}
}

@misc{wen2026verifierfreerlllmsintrinsic,
  title         = {Verifier-Free RL for LLMs via Intrinsic Gradient-Norm Reward},
  author        = {Xuexiang Wen and Hang Yu and Linchao Zhu and Gaoang Wang},
  year          = {2026},
  eprint        = {2605.09920},
  archiveprefix = {arXiv},
  primaryclass  = {cs.LG},
  url           = {https://arxiv.org/abs/2605.09920}
}

@misc{wu2024metarewarding,
  title         = {Meta-Rewarding Language Models: Self-Improving Alignment with LLM-as-a-Meta-Judge},
  author        = {Tianhao Wu and Weizhe Yuan and Olga Golovneva and Jing Xu and Yuandong Tian and Jiantao Jiao and Jason Weston and Sainbayar Sukhbaatar},
  year          = {2024},
  eprint        = {2407.19594},
  archiveprefix = {arXiv},
  primaryclass  = {cs.CL},
  url           = {https://arxiv.org/abs/2407.19594}
}

@misc{wu2025spine,
  title         = {SPINE: Token-Selective Test-Time Reinforcement Learning with Entropy-Band Regularization},
  author        = {Jianghao Wu and Yasmeen George and Jin Ye and Yicheng Wu and Daniel F. Schmidt and Jianfei Cai},
  year          = {2025},
  eprint        = {2511.17938},
  archiveprefix = {arXiv},
  primaryclass  = {cs.CL},
  url           = {https://arxiv.org/abs/2511.17938}
}

@misc{wu2026modelsjudgethemselvesunsupervised,
  title         = {When Models Judge Themselves: Unsupervised Self-Evolution for Multimodal Reasoning},
  author        = {Zhengxian Wu and Kai Shi and Chuanrui Zhang and Zirui Liao and Jun Yang and Ni Yang and Qiuying Peng and Luyuan Zhang and Hangrui Xu and Tianhuang Su and Zhenyu Yang and Haonan Lu and Haoqian Wang},
  year          = {2026},
  eprint        = {2603.21289},
  archiveprefix = {arXiv},
  primaryclass  = {cs.CV},
  url           = {https://arxiv.org/abs/2603.21289}
}

@misc{xie2026sslr1selfsupervisedvisualreinforcement,
  title         = {SSL-R1: Self-Supervised Visual Reinforcement Post-Training for Multimodal Large Language Models},
  author        = {Jiahao Xie and Alessio Tonioni and Nathalie Rauschmayr and Federico Tombari and Bernt Schiele},
  year          = {2026},
  eprint        = {2604.20705},
  archiveprefix = {arXiv},
  primaryclass  = {cs.CV},
  url           = {https://arxiv.org/abs/2604.20705}
}

@misc{xing2025compass,
  title         = {Rewarding the Journey, Not Just the Destination: A Composite Path and Answer Self-Scoring Reward Mechanism for Test-Time Reinforcement Learning},
  author        = {Jingyu Xing and Chenwei Tang and Xinyu Liu and Deng Xiong and Shudong Huang and Wei Ju and Jiancheng Lv and Ziyue Qiao},
  year          = {2025},
  eprint        = {2510.17923},
  archiveprefix = {arXiv},
  primaryclass  = {cs.LG},
  url           = {https://arxiv.org/abs/2510.17923}
}

@inproceedings{xiong2024longcontext,
  title     = {Effective Long-Context Scaling of Foundation Models},
  author    = {Wenhan Xiong and Jingyu Liu and Igor Molybog and Hejia Zhang and Prajjwal Bhargava and Rui Hou and Louis Martin and Rashi Rungta and Karthik Abinav Sankararaman and Barlas Oguz and Madian Khabsa and Han Fang and Yashar Mehdad and Sharan Narang and Kshitiz Malik and Angela Fan and Shruti Bhosale and Sergey Edunov and Mike Lewis and Sinong Wang and Hao Ma},
  booktitle = {Proceedings of the 2024 Conference of the North American Chapter of the Association for Computational Linguistics: Human Language Technologies},
  pages     = {4643--4663},
  year      = {2024},
  doi       = {10.18653/v1/2024.naacl-long.260},
  url       = {https://aclanthology.org/2024.naacl-long.260/}
}

@misc{xu2025genius,
  title         = {Genius: A Generalizable and Purely Unsupervised Self-Training Framework For Advanced Reasoning},
  author        = {Fangzhi Xu and Hang Yan and Chang Ma and Haiteng Zhao and Qiushi Sun and Kanzhi Cheng and Junxian He and Jun Liu and Zhiyong Wu},
  year          = {2025},
  eprint        = {2504.08672},
  archiveprefix = {arXiv},
  primaryclass  = {cs.CL},
  url           = {https://arxiv.org/abs/2504.08672}
}

@misc{xu2025reasoningmodels,
  title         = {Towards Large Reasoning Models: A Survey of Reinforced Reasoning with Large Language Models},
  author        = {Fengli Xu and Qianyue Hao and Zefang Zong and Jingwei Wang and Yunke Zhang and Jingyi Wang and Xiaochong Lan and Jiahui Gong and Tianjian Ouyang and Fanjin Meng and Chenyang Shao and Yuwei Yan and Qinglong Yang and Yiwen Song and Sijian Ren and Xinyuan Hu and Yu Li and Jie Feng and Chen Gao and Yong Li},
  year          = {2025},
  eprint        = {2501.09686},
  archiveprefix = {arXiv},
  primaryclass  = {cs.AI},
  url           = {https://arxiv.org/abs/2501.09686}
}

@misc{xu2025you,
  title         = {You only need 4 extra tokens: Synergistic Test-time Adaptation for LLMs},
  author        = {Yijie Xu and Huizai Yao and Zhiyu Guo and Pengteng Li and Aiwei Liu and Xuming Hu and Weiyu Guo and Hui Xiong},
  year          = {2025},
  eprint        = {2510.10223},
  archiveprefix = {arXiv},
  primaryclass  = {cs.CL},
  url           = {https://arxiv.org/abs/2510.10223}
}

@misc{xu2026uldtta,
  title         = {Unsupervised Layer-Wise Dynamic Test Time Adaptation for LLMs},
  author        = {Longhuan Xu and Cunjian Chen and Feng Yin},
  year          = {2026},
  eprint        = {2602.09719},
  archiveprefix = {arXiv},
  primaryclass  = {cs.CL},
  url           = {https://arxiv.org/abs/2602.09719}
}

@misc{yan2026scrl,
  title         = {What If Consensus Lies? Selective-Complementary Reinforcement Learning at Test Time},
  author        = {Dong Yan and Jian Liang and Yanbo Wang and Shuo Lu and Ran He and Tieniu Tan},
  year          = {2026},
  eprint        = {2603.19880},
  archiveprefix = {arXiv},
  primaryclass  = {cs.LG},
  url           = {https://arxiv.org/abs/2603.19880}
}

@misc{yang2026ttcs,
  title         = {TTCS: Test-Time Curriculum Synthesis for Self-Evolving},
  author        = {Chengyi Yang and Zhishang Xiang and Yunbo Tang and Zongpei Teng and Chengsong Huang and Fei Long and Yuhan Liu and Jinsong Su},
  year          = {2026},
  eprint        = {2601.22628},
  archiveprefix = {arXiv},
  primaryclass  = {cs.LG},
  url           = {https://arxiv.org/abs/2601.22628}
}

@misc{yu2026stabilizingunsupervisedselfevolutionmllms,
  title         = {Stabilizing Unsupervised Self-Evolution of MLLMs via Continuous Softened Retracing reSampling},
  author        = {Yunyao Yu and Zhengxian Wu and Zhuohong Chen and Hangrui Xu and Zirui Liao and Xiangwen Deng and Zhifang Liu and Senyuan Shi and Haoqian Wang},
  year          = {2026},
  eprint        = {2604.03647},
  archiveprefix = {arXiv},
  primaryclass  = {cs.CV},
  url           = {https://arxiv.org/abs/2604.03647}
}

@inproceedings{yuan2024selfrewarding,
  title     = {Self-Rewarding Language Models},
  author    = {Yuan, Weizhe and Pang, Richard Yuanzhe and Cho, Kyunghyun and Li, Xian and Sukhbaatar, Sainbayar and Xu, Jing and Weston, Jason E},
  booktitle = {Proceedings of the 41st International Conference on Machine Learning},
  pages     = {57905--57923},
  year      = {2024},
  editor    = {Salakhutdinov, Ruslan and Kolter, Zico and Heller, Katherine and Weller, Adrian and Oliver, Nuria and Scarlett, Jonathan and Berkenkamp, Felix},
  volume    = {235},
  series    = {Proceedings of Machine Learning Research},
  month     = {21--27 Jul},
  publisher = {PMLR},
  url       = {https://proceedings.mlr.press/v235/yuan24d.html}
}

@misc{yuan2025rlccf,
  title         = {Wisdom of the Crowd: Reinforcement Learning from Coevolutionary Collective Feedback},
  author        = {Wenzhen Yuan and Shengji Tang and Weihao Lin and Jiacheng Ruan and Ganqu Cui and Bo Zhang and Tao Chen and Ting Liu and Yuzhuo Fu and Peng Ye and Lei Bai},
  year          = {2025},
  eprint        = {2508.12338},
  archiveprefix = {arXiv},
  primaryclass  = {cs.AI},
  url           = {https://arxiv.org/abs/2508.12338}
}

@inproceedings{zelikman2024quietstar,
  title         = {Quiet-{STaR}: Language Models Can Teach Themselves to Think Before Speaking},
  author        = {Eric Zelikman and Georges Harik and Yijia Shao and Varuna Jayasiri and Nick Haber and Noah D. Goodman},
  booktitle     = {First Conference on Language Modeling},
  year          = {2024},
  eprint        = {2403.09629},
  archiveprefix = {arXiv},
  url           = {https://openreview.net/forum?id=oRXPiSOGH9}
}

@inproceedings{zeng2025kbalign,
  title     = {{KBA}lign: Efficient Self Adaptation on Specific Textual Knowledge Bases},
  author    = {Zeng, Zheni  and
               Chen, Yuxuan  and
               Yu, Shi  and
               Wang, Ruobing  and
               Yan, Yukun  and
               Liu, Zhenghao  and
               Wang, Shuo  and
               Han, Xu  and
               Liu, Zhiyuan  and
               Sun, Maosong},
  editor    = {Christodoulopoulos, Christos  and
               Chakraborty, Tanmoy  and
               Rose, Carolyn  and
               Peng, Violet},
  booktitle = {Findings of the Association for Computational Linguistics: EMNLP 2025},
  month     = nov,
  year      = {2025},
  address   = {Suzhou, China},
  publisher = {Association for Computational Linguistics},
  url       = {https://aclanthology.org/2025.findings-emnlp.728/},
  doi       = {10.18653/v1/2025.findings-emnlp.728},
  pages     = {13519--13532},
  isbn      = {979-8-89176-335-7}
}

@misc{zhang2025covo,
  title         = {Consistent Paths Lead to Truth: Self-Rewarding Reinforcement Learning for LLM Reasoning},
  author        = {Kongcheng Zhang and Qi Yao and Shunyu Liu and Yingjie Wang and Baisheng Lai and Jieping Ye and Mingli Song and Dacheng Tao},
  year          = {2025},
  eprint        = {2506.08745},
  archiveprefix = {arXiv},
  primaryclass  = {cs.AI},
  url           = {https://arxiv.org/abs/2506.08745}
}

@misc{zhang2025lepa,
  title         = {Learning to Plan Before Answering: Self-Teaching LLMs to Learn Abstract Plans for Problem Solving},
  author        = {Jin Zhang and Flood Sung and Zhilin Yang and Yang Gao and Chongjie Zhang},
  year          = {2025},
  eprint        = {2505.00031},
  archiveprefix = {arXiv},
  primaryclass  = {cs.CL},
  url           = {https://arxiv.org/abs/2505.00031}
}

@misc{zhang2025nofreelunch,
  title         = {No Free Lunch: Rethinking Internal Feedback for LLM Reasoning},
  author        = {Yanzhi Zhang and Zhaoxi Zhang and Haoxiang Guan and Yilin Cheng and Yitong Duan and Chen Wang and Yue Wang and Shuxin Zheng and Jiyan He},
  year          = {2025},
  eprint        = {2506.17219},
  archiveprefix = {arXiv},
  primaryclass  = {cs.LG},
  url           = {https://arxiv.org/abs/2506.17219}
}

@misc{zhang2025rightquestion,
  title         = {Right Question is Already Half the Answer: Fully Unsupervised LLM Reasoning Incentivization},
  author        = {Qingyang Zhang and Haitao Wu and Changqing Zhang and Peilin Zhao and Yatao Bian},
  year          = {2025},
  eprint        = {2504.05812},
  archiveprefix = {arXiv},
  primaryclass  = {cs.LG},
  url           = {https://arxiv.org/abs/2504.05812},
  note          = {Ongoing work. First released on April 8, 2025}
}

@inproceedings{zhang2025selftuning,
  title     = {Self-Tuning: Instructing {LLM}s to Effectively Acquire New Knowledge through Self-Teaching},
  author    = {Zhang, Xiaoying  and
               Peng, Baolin  and
               Tian, Ye  and
               Zhou, Jingyan  and
               Zhang, Yipeng  and
               Mi, Haitao  and
               Meng, Helen M.},
  editor    = {Che, Wanxiang  and
               Nabende, Joyce  and
               Shutova, Ekaterina  and
               Pilehvar, Mohammad Taher},
  booktitle = {Findings of the Association for Computational Linguistics: ACL 2025},
  month     = jul,
  year      = {2025},
  address   = {Vienna, Austria},
  publisher = {Association for Computational Linguistics},
  url       = {https://aclanthology.org/2025.findings-acl.297/},
  doi       = {10.18653/v1/2025.findings-acl.297},
  pages     = {5688--5724},
  isbn      = {979-8-89176-256-5}
}

@inproceedings{zhang2026corewarding,
  title     = {Co-rewarding: Stable Self-supervised {RL} for Eliciting Reasoning in Large Language Models},
  author    = {Zizhuo Zhang and Jianing Zhu and Xinmu Ge and Zihua Zhao and Zhanke Zhou and Xuan Li and Xiao Feng and Jiangchao Yao and Bo Han},
  booktitle = {The Fourteenth International Conference on Learning Representations},
  year      = {2026},
  url       = {https://openreview.net/forum?id=fDk95XPsCU}
}

@misc{zhang2026silencejudgereinforcementlearning,
  title         = {Silence the Judge: Reinforcement Learning with Self-Verifier via Latent Geometric Clustering},
  author        = {Nonghai Zhang and Weitao Ma and Zhanyu Ma and Jun Xu and Jiuchong Gao and Jinghua Hao and Renqing He and Jingwen Xu},
  year          = {2026},
  eprint        = {2601.08427},
  archiveprefix = {arXiv},
  primaryclass  = {cs.CL},
  url           = {https://arxiv.org/abs/2601.08427}
}

@misc{zhao2025absolutezero,
  title         = {Absolute Zero: Reinforced Self-play Reasoning with Zero Data},
  author        = {Andrew Zhao and Yiran Wu and Yang Yue and Tong Wu and Quentin Xu and Yang Yue and Matthieu Lin and Shenzhi Wang and Qingyun Wu and Zilong Zheng and Gao Huang},
  year          = {2025},
  eprint        = {2505.03335},
  archiveprefix = {arXiv},
  primaryclass  = {cs.LG},
  url           = {https://arxiv.org/abs/2505.03335}
}

@inproceedings{zhao2025intuitor,
  title     = {Learning to Reason without External Rewards},
  author    = {Xuandong Zhao and Zhewei Kang and Aosong Feng and Sergey Levine and Dawn Song},
  booktitle = {The Fourteenth International Conference on Learning Representations},
  year      = {2026},
  url       = {https://openreview.net/forum?id=OU9nFEYR2M}
}

@misc{zhao2026echo,
  title         = {ECHO: Entropy-Confidence Hybrid Optimization for Test-Time Reinforcement Learning},
  author        = {Chu Zhao and Enneng Yang and Yuting Liu and Jianzhe Zhao and Guibing Guo},
  year          = {2026},
  eprint        = {2602.02150},
  archiveprefix = {arXiv},
  primaryclass  = {cs.LG},
  url           = {https://arxiv.org/abs/2602.02150}
}

@misc{zhou2024csr,
  title         = {Calibrated Self-Rewarding Vision Language Models},
  author        = {Yiyang Zhou and Zhiyuan Fan and Dongjie Cheng and Sihan Yang and Zhaorun Chen and Chenhang Cui and Xiyao Wang and Yun Li and Linjun Zhang and Huaxiu Yao},
  year          = {2024},
  eprint        = {2405.14622},
  archiveprefix = {arXiv},
  primaryclass  = {cs.LG},
  url           = {https://arxiv.org/abs/2405.14622}
}

@misc{zhou2025evolrl,
  title         = {Evolving Language Models without Labels: Majority Drives Selection, Novelty Promotes Variation},
  author        = {Yujun Zhou and Zhenwen Liang and Haolin Liu and Wenhao Yu and Kishan Panaganti and Linfeng Song and Dian Yu and Xiangliang Zhang and Haitao Mi and Dong Yu},
  year          = {2025},
  eprint        = {2509.15194},
  archiveprefix = {arXiv},
  primaryclass  = {cs.LG},
  url           = {https://arxiv.org/abs/2509.15194}
}

@inproceedings{zuo2025ttrl,
  title     = {{TTRL}: Test-Time Reinforcement Learning},
  author    = {Yuxin Zuo and Kaiyan Zhang and Li Sheng and Shang Qu and Ganqu Cui and Xuekai Zhu and Haozhan Li and Yuchen Zhang and Xinwei Long and Ermo Hua and Biqing Qi and Youbang Sun and Zhiyuan Ma and Lifan Yuan and Ning Ding and Bowen Zhou},
  booktitle = {Advances in Neural Information Processing Systems},
  volume    = {38},
  year      = {2025},
  doi       = {10.52202/085713-4376},
  url       = {https://proceedings.neurips.cc/paper_files/paper/2025/hash/be690ea16f005c174f6c4102a5970e67-Abstract-Conference.html}
}

@misc{liang2024internalconsistencyselffeedbacklarge,
      title={Internal Consistency and Self-Feedback in Large Language Models: A Survey},
      author={Xun Liang and Shichao Song and Zifan Zheng and Hanyu Wang and Qingchen Yu and Xunkai Li and Rong-Hua Li and Yi Wang and Zhonghao Wang and Feiyu Xiong and Zhiyu Li},
      year={2024},
      eprint={2407.14507},
      archivePrefix={arXiv},
      primaryClass={cs.CL},
      url={https://arxiv.org/abs/2407.14507},
}

@misc{tie2025surveyposttraininglargelanguage,
      title={A Survey on Post-training of Large Language Models},
      author={Guiyao Tie and Zeli Zhao and Dingjie Song and Fuyang Wei and Rong Zhou and Yurou Dai and Wen Yin and Zhejian Yang and Jiangyue Yan and Yao Su and Zhenhan Dai and Yifeng Xie and Yihan Cao and Lichao Sun and Pan Zhou and Lifang He and Hechang Chen and Yu Zhang and Qingsong Wen and Tianming Liu and Neil Zhenqiang Gong and Jiliang Tang and Caiming Xiong and Heng Ji and Philip S. Yu and Jianfeng Gao},
      year={2025},
      eprint={2503.06072},
      archivePrefix={arXiv},
      primaryClass={cs.CL},
      url={https://arxiv.org/abs/2503.06072},
}

@misc{wu2025sailingstarssurveyreward,
      title={Sailing by the Stars: A Survey on Reward Models and Learning Strategies for Learning from Rewards},
      author={Xiaobao Wu},
      year={2025},
      eprint={2505.02686},
      archivePrefix={arXiv},
      primaryClass={cs.CL},
      url={https://arxiv.org/abs/2505.02686},
}

@misc{deng2025selfimprovementmultimodallargelanguage,
      title={Self-Improvement in Multimodal Large Language Models: A Survey},
      author={Shijian Deng and Kai Wang and Tianyu Yang and Harsh Singh and Yapeng Tian},
      year={2025},
      eprint={2510.02665},
      archivePrefix={arXiv},
      primaryClass={cs.CL},
      url={https://arxiv.org/abs/2510.02665},
}

@misc{yang2026selfimprovementlargelanguagemodels,
      title={Self-Improvement of Large Language Models: A Technical Overview and Future Outlook},
      author={Haoyan Yang and Mario Xerri and Solha Park and Huajian Zhang and Yiyang Feng and Sai Akhil Kogilathota and Jiawei Zhou},
      year={2026},
      eprint={2603.25681},
      archivePrefix={arXiv},
      primaryClass={cs.CL},
      url={https://arxiv.org/abs/2603.25681},
}

@inproceedings{dai2025seper,
    title={SePer: Measure Retrieval Utility Through The Lens Of Semantic Perplexity Reduction},
    author={Dai, Lu and Xu, Yijie and Ye, Jinhui and Liu, Hao and Xiong, Hui},
    booktitle={International Conference on Learning Representations (ICLR)},
    year={2025}
}

@inproceedings{li2025seetrek,
  title={See\&Trek: Training-Free Spatial Prompting for Multimodal Large Language Model},
  author={Pengteng Li and Pinhao Song and Wuyang Li and Huizai Yao and Weiyu Guo and Yijie Xu and Dugang Liu and Hui Xiong},
  booktitle={The Thirty-ninth Annual Conference on Neural Information Processing Systems},
  year={2025},
  url={https://openreview.net/forum?id=2exr4mlbx1}
}

\clearpage
\appendix

\section*{Appendix Contents}
\small
\noindent
\begin{tabularx}{\columnwidth}{@{}lXr@{}}
  \textbf{A} & \hyperref[app:protocol]{Survey Protocol and Screening Details}
  & \pageref{app:protocol} \\[0.3em]
  \textbf{B} & \hyperref[app:synthesis-evidence]{Representative Evidence and Task-Structure Audit}
  & \pageref{app:synthesis-evidence} \\[0.3em]
  \textbf{C} & \hyperref[app:inventory]{Full Method Inventory}
  & \pageref{app:inventory} \\[0.3em]
  \textbf{D} & \hyperref[app:adjacent]{Boundary Decisions}
  & \pageref{app:adjacent} \\[0.3em]
  \textbf{E} & \hyperref[app:survey-comparison]{Comparison with Existing Surveys}
  & \pageref{app:survey-comparison} \\[0.3em]
  \textbf{F} & \hyperref[app:ai-assistants]{Use of AI Assistants}
  & \pageref{app:ai-assistants} \\
\end{tabularx}
\normalsize

\section{Survey Protocol and Screening Details}
\label{app:protocol}

\subsection{Search Sources and Query Strings}
We searched ACL Anthology, arXiv, Semantic Scholar, and Google
Scholar for work dated January~2023--May~2026. The following frozen
templates make the search reconstructable; each quoted mechanism
phrase was run separately to avoid engine-specific Boolean limits.

\begin{itemize}[leftmargin=*,nosep]
  \item \textbf{ACL Anthology search:}
    \texttt{"<mechanism>" ("large language model" OR LLM OR multimodal)}.
  \item \textbf{arXiv API:}
    \texttt{(ti:"<mechanism>" OR abs:"<mechanism>") AND
      (all:"large language model" OR all:"multimodal large language
    model")}, with submitted-date bounds 2023-01-01 and 2026-05-31.
  \item \textbf{Semantic Scholar API:}
    \texttt{query="<mechanism> large language model"},
    \texttt{year=2023-2026}, \texttt{fieldsOfStudy=Computer Science}.
  \item \textbf{Google Scholar:}
    \texttt{"<mechanism>" ("large language model" OR LLM OR MLLM)
    -survey}, with a custom 2023--2026 year range.
\end{itemize}

The mechanism list was: \emph{unsupervised post-training},
\emph{self-improvement}, \emph{self-rewarding},
\emph{self-training}, \emph{test-time training}, \emph{test-time
adaptation}, \emph{test-time reinforcement learning},
\emph{internal reward}, \emph{self-consistency}, \emph{majority
vote}, \emph{intrinsic reward}, and \emph{evaluator-driven RL}.
Forward and backward snowballing expanded the four strands in
\S\ref{sec:definitions}.

\subsection{Inclusion and Exclusion Criteria}
\textbf{Inclusion}: a candidate is included as strict UPT iff it
satisfies all four boundary checks \bcref{B1}--\bcref{B4} of
\S\ref{sec:definitions}, has a verifiable algorithmic description
in a paper, preprint, or extended technical report, and operates on
foundation-scale text or multimodal models. Methods on smaller
classifiers, image-only encoders predating LLM/MLLM
self-improvement, or pure decoding-time tricks without an explicit
update are not included as strict UPT. Tool-grounded,
verifier-grounded, seed-supervised, stronger-teacher, and
external-evaluator methods are kept as adjacent methods.
\textbf{Exclusion}: surveys, benchmarks-only papers, hardware/system
papers, and unrelated domain-specific applications without a
post-training contribution.

\subsection{Screening Flow}
Each candidate paper was reviewed with a structured note recording:
update target (parameters, adapters, memories, persistent local
state, none); signal source (model samples, internal aggregates,
internal evaluator, external verifier, external label); whether any
seed, tool, or stronger-teacher signal entered the update loop; the
internal update object against which the gradient is actually
computed; and the timing regime (offline corpus, full-cohort
  transductive, few-sample target, streaming continual, test-time
instance, within-sequence, or no-update inference-time).
Boundary cases receive a source-section recheck and a recorded
assignment rationale in Appendix~\ref{app:reliability}.

After deduplication, the inventory contains \textbf{94 method
records from 91 papers}: \textbf{80 strict rows from 78 papers},
\textbf{8 adjacent rows}, and \textbf{6 prose-only boundary or
antecedent records}. Multiple algorithmic variants in one paper are
separate method rows.

\clearpage
\begin{strip}
  \centering
  \scriptsize
  \setlength{\tabcolsep}{3.5pt}
  \renewcommand{\arraystretch}{1.10}
  \begin{tabularx}{\textwidth}{@{}lLLLL@{}}
    \toprule
    \textbf{Family / method} &
    \textbf{Backbone and evaluation} &
    \textbf{Internal update signal} &
    \textbf{Originally reported result} &
    \textbf{Interpretation and task structure} \\
    \midrule
    I / LangAdapt CPT~\citep{elhady2025languageadapt} &
    Llama~2 7B; Basque LM PPL, downstream accuracy, Copain &
    Next-token NLL on unlabeled Basque and English text &
    PPL $23.64\!\rightarrow\!3.35$; downstream $27.43\!\rightarrow\!34.14$; Copain $44.67\!\rightarrow\!43.43$ &
    Large perplexity and downstream gains coexist with task-specific variation, separating language adaptation from multiple-choice evaluation. \\
    II / TTRL~\citep{zuo2025ttrl} &
    Qwen2.5-Math-7B; pass@1 on AIME~2024, MATH-500, GPQA-Diamond &
    Same-model answer majority as GRPO reward &
    AIME $12.9\!\rightarrow\!40.2$; MATH-500 $46.7\!\rightarrow\!83.4$; GPQA $29.1\!\rightarrow\!27.7$ &
    Transductive adaptation yields large benchmark-specific gains; the GPQA result shows that the update does not automatically transfer across evaluation distributions. \\
    III / Quiet-STaR~\citep{zelikman2024quietstar} &
    Mistral 7B; zero-shot GSM8K and CommonsenseQA &
    Thoughts selected by improvement in future-token prediction &
    GSM8K $5.9\!\rightarrow\!10.9$; CommonsenseQA $36.3\!\rightarrow\!47.2$ &
    Zero-shot gains arise from future-token-trained thoughts, with thought and lookahead lengths defining the operating point. \\
    IV / \textsc{CoNL}~\citep{sui2026conl} &
    Qwen3-8B; pass@1 on AIME~2024 and DeepMath &
    Same-policy multi-agent critiques and rankings as diagnostic reward &
    AIME $60.0\!\rightarrow\!76.5$; DeepMath $70.5\!\rightarrow\!87.1$ &
    Same-policy multi-agent evaluation yields strong reasoning gains without an external judge; rollout cost and open-ended transfer define the next evaluation axes. \\
    \bottomrule
  \end{tabularx}
  \captionsetup{hypcap=false}
  \captionof{table}{Representative results, one method per family. The table
    preserves each source paper's backbone, data, budget, and checkpoint
  rules to expose mechanism-specific evidence and task conditions.}
  \label{tab:representative-evidence}
\end{strip}

\begin{strip}
  \centering
  \scriptsize
  \setlength{\tabcolsep}{4pt}
  \renewcommand{\arraystretch}{1.08}
  \begin{tabularx}{\textwidth}{@{}lLLL@{}}
    \toprule
    \textbf{Task-structure prior} &
    \textbf{What it contributes} &
    \textbf{Representative exposure} &
    \textbf{Boundary interpretation} \\
    \midrule
    Answer extractor / canonicalizer &
    An equivalence relation over free-form strings; easier vote counting &
    Math-answer consensus in TTRL, ETTRL, and related Family~II methods &
    Supplies an equivalence relation without correctness supervision; sensitivity should be evaluated across alternative normalizers. \\
    Finite answer alphabet &
    A small support for clustering or voting &
    Multiple-choice reasoning and classification evaluations &
    Supplies a compact comparison space; open-ended reformulations test whether the gain extends beyond that structure. \\
    Code signature vs. execution &
    A signature constrains output form; execution returns a correctness-bearing verdict &
    Code-oriented self-training and adjacent executor-based methods &
    Signature alone is a prior; tests, interpreters, and environment rewards fail B3. \\
    Full-cohort transduction &
    Visibility of the target input distribution before final prediction &
    TTRL, MM-UPT, and other cohort-visible test-time updates &
    Provides target-distribution visibility and should be reported separately from held-out generalization. \\
    Open-ended output space &
    No canonical equality test; comparison requires generated targets or an evaluator &
    G-Zero, \textsc{CoNL}, dialogue, summarization, and creative generation &
    Exposes the regime where generated targets and internal evaluators provide the comparison signal. \\
    \bottomrule
  \end{tabularx}
  \captionsetup{hypcap=false}
  \captionof{table}{Task structure is audited independently from explicit
    supervision. A prior can make an internal signal more informative
    without specifying the correct answer; an executed verifier crosses
  the strict boundary.}
  \label{tab:task-prior-audit}
\end{strip}

\section{Representative Evidence and Task-Structure Audit}
\label{app:synthesis-evidence}
Tables~\ref{tab:representative-evidence} and
\ref{tab:task-prior-audit} provide the empirical and task-structure
detail referenced by the main-text synthesis. The first preserves
each source paper's reported setup; the second records structural
priors separately from correctness-bearing supervision.

\clearpage
\begin{figure*}[p]
  \begin{minipage}{\textwidth}
  \centering
  \scriptsize
  \setlength{\tabcolsep}{4pt}
  \renewcommand{\arraystretch}{1.12}
  \begin{tabular}{@{}lrrrrrrrrr@{}}
    \toprule
    & \multicolumn{6}{c}{\textbf{Primary timing regime}}
    & \multicolumn{2}{c}{\textbf{Update target}} & \\
    \cmidrule(lr){2-7}\cmidrule(lr){8-9}
    \textbf{Family} & \textbf{Off.} & \textbf{Coh.} & \textbf{Few}
    & \textbf{Str.} & \textbf{Inst.} & \textbf{Seq.}
    & \textbf{Param.} & \textbf{Local} & \textbf{Total} \\
    \midrule
    I   & 17 & 0 & 1 & 1 & 6 & 1 & 22 & 4 & 26 \\
    II  & 15 & 7 & 0 & 0 & 0 & 0 & 22 & 0 & 22 \\
    III & 18 & 3 & 0 & 0 & 1 & 0 & 21 & 1 & 22 \\
    IV  & 9  & 1 & 0 & 0 & 0 & 0 & 10 & 0 & 10 \\
    \midrule
    \textbf{Total} & \textbf{59} & \textbf{11} & \textbf{1}
    & \textbf{1} & \textbf{7} & \textbf{1}
    & \textbf{75} & \textbf{5} & \textbf{80} \\
    \bottomrule
  \end{tabular}
  \captionsetup{hypcap=false}
  \captionof{table}{Cross-section of the strict inventory by family, primary
    timing regime, and update target. Family~I includes the
    \textsc{SUDER} bridge row. Off.: offline corpus; Coh.: full-cohort
    transductive; Few: few-sample target; Str.: streaming continual;
    Inst.: test-time instance; Seq.: within-sequence; Local:
  sample-local state. Each method record contributes once.}
  \label{tab:inventory-crosstab}

  \vspace{0.75em}
  \normalsize
  \begin{adjustbox}{max totalsize={0.99\textwidth}{0.57\textheight},center}
    \usebox{\taxonomytreeboxFULL}
  \end{adjustbox}
  \vspace{-0.5em}
  \captionsetup{hypcap=false}
  \captionof{figure}{Full update-object taxonomy. Each of the 80 strict UPT
    methods appears once under its primary family and sub-class;
    $^{\ddagger}$ marks the Family~I/IV bridge cases
    (\textsc{SUDER} and \textsc{GvU}). The dashed adjacent branch
  places eight neighboring methods by the boundary they cross.}
  \vspace{-1em}
  \label{fig:taxonomy-extended}
  \end{minipage}
\end{figure*}

\clearpage
\section{Full Method Inventory}
\label{app:inventory}

Figure~\ref{fig:taxonomy-extended} maps all 80 strict methods by
family, sub-class, and update object. Tables~\ref{tab:family-i}--
\ref{tab:family-iv} compare method attributes; the tree provides the
complete hierarchy in a single view.

\subsection{Family, Timing, and Update Target}
The
\href{https://github.com/yeahjack/awesome-unsupervised-post-training}{companion inventory}
provides machine-readable method records and per-paper rationales.
Table~\ref{tab:inventory-crosstab} aggregates its 80 strict rows
along the timing and update-target axes used in the main text.

Offline updates dominate all four families (59/80). Full-cohort
transduction is concentrated in Family~II (7/11), whereas all five
sample-local-state methods occur in Families~I and~III. The
cross-tabulation therefore connects the update-object taxonomy to
the deployment regimes in \S\ref{sec:timing}.

\section{Boundary Decisions}
\label{app:adjacent}
\label{app:reliability}

Adjacent methods are organized by the boundary check they fail.
The cases below cover the recurring ambiguities; stronger-teacher
distillation fails~\bcref{B2}--\bcref{B3} directly.

\subsection{Representative Adjacent Cases}
\textbf{No-update inference-time optimization.}
\textsc{EM-INF}~\citep{agarwal2025unreasonable} performs
inference-time entropy descent over logits or hidden states
without modifying parameters, adapters, memories, or persistent
local state, failing~\bcref{B1}.
Training-free multimodal prompting provides the same boundary test:
\textsc{See\&Trek}~\citep{li2025seetrek} changes spatial
prompt construction without updating model parameters or persistent
state. Likewise, \textsc{SePer}~\citep{dai2025seper}
uses semantic-perplexity reduction to measure retrieval utility;
an internal model statistic does not satisfy~\bcref{B1} unless it
drives an explicit update.

\textbf{Verifier- or tool-assisted self-training.}
\textsc{T$^3$RL}~\citep{liao2026t3rl} pairs majority-vote selection
with code-interpreter verification; \textsc{Absolute
Zero}~\citep{zhao2025absolutezero} closes a propose-solve loop with
a code executor as the truth oracle. Both fail~\bcref{B3}. They are
important neighboring evidence for false-popular collapse fixes
(\S\ref{sec:open-problems}).

\textbf{Human- or seed-supervised bootstrapping.}
\textsc{Self-Instruct}~\citep{wang2023selfinstruct} and
instruction-backtranslation
pipelines~\citep{li2024instructionbacktranslation} bootstrap from
human-written seeds, failing~\bcref{B3} at the seed stage. They are
treated as precursors of self-generated target bootstrapping rather
than strict UPT.

\textbf{External reward or evaluator methods.}
The full \textsc{CSR}~\citep{zhou2024csr} system includes a
CLIP-derived visual-relevance term in its reward. Because the
evaluator is not derived from the same model lineage,
\textsc{CSR}~fails~\bcref{B4}. It is routed to adjacent unless an
internal-only variant is analyzed separately.

\subsection{Family II vs.\ Family III}
\textsc{LRM~Self-Train}~\citep{shi2025lrmselftrain} uses majority
vote to filter candidate solutions before SFT on the survivors.
Because the gradient is computed against the kept solutions
(self-generated targets), not against the consensus statistic
itself, the update-object rule assigns it to Family~III. The
same rule places \textsc{TTRL}, \textsc{RoiRL}, and methods whose
gradient is computed directly against
$r=\mathbf{1}[y=\mathrm{maj}]$ in Family~II.

\subsection{Family III vs.\ Family IV}
\textsc{Confident~ST}~\citep{wang2025confidentreasoning} and
\textsc{RLSF}~\citep{vanniekerk2025rlsf} use a self-confidence or
self-rated score on candidate trajectories. When the score acts as
a selection mask before SFT or DPO, the gradient is computed
against the kept generations (Family~III). When the score itself
appears as the scalar reward in PG, the gradient is computed
through the evaluator (Family~IV). The update-object rule assigns
each method according to its actual gradient path; the companion
inventory records the mapping.

\subsection{Strict UPT vs.\ Adjacent Methods}
\textsc{ECHO} and \textsc{SPINE} use multi-sample consensus as a
selection mask while a separate intrinsic term shapes advantages or
selects gradient-receiving tokens. Their consensus reward places
them in Family~II, with the intrinsic term acting as a within-family
modulation. \textsc{EM-INF} and the full
\textsc{CSR} are routed to adjacent for the reasons in
Appendix~\ref{app:adjacent}.

\clearpage
\begin{strip}
  \centering
  \footnotesize
  \setlength{\tabcolsep}{3.5pt}
  \renewcommand{\arraystretch}{1.10}
  \begin{tabularx}{\textwidth}{@{}L l cccc@{}}
    \toprule
    \textbf{Survey} & \textbf{Venue} &
    \textbf{Real} & \textbf{No external} & \textbf{Update-object} &
    \textbf{MLLM} \\
    & & \textbf{update} & \textbf{signal} & \textbf{taxonomy} &
    \textbf{coverage} \\
    \midrule
    \citet{liang2025ttasurvey} (\textit{TTA}) &
    arXiv 2023 & \hbfull & \hbhalf & \hbnone & \hbnone \\
    \citet{tao2024selfevolution} (\textit{Self-Evolution}) &
    arXiv 2024 & \hbhalf & \hbhalf & \hbnone & \hbhalf \\
    \citet{liang2024internalconsistencyselffeedbacklarge}
    (\textit{Internal Consistency}) &
    arXiv 2024 & \hbhalf & \hbhalf & \hbnone & \hbnone \\
    \citet{xu2025reasoningmodels} (\textit{Reinforced Reasoning}) &
    arXiv 2025 & \hbhalf & \hbnone & \hbnone & \hbnone \\
    \citet{kumar2025posttraining} (\textit{LLM Post-Training}) &
    arXiv 2025 & \hbhalf & \hbnone & \hbnone & \hbhalf \\
    \citet{chen2025longcotsurvey} (\textit{Long CoT}) &
    arXiv 2025 & \hbhalf & \hbnone & \hbnone & \hbfull \\
    \citet{tie2025surveyposttraininglargelanguage}
    (\textit{Alignment to Reasoning}) &
    arXiv 2025 & \hbhalf & \hbnone & \hbnone & \hbhalf \\
    \citet{wu2025sailingstarssurveyreward}
    (\textit{Learning from Rewards}) &
    arXiv 2025 & \hbhalf & \hbhalf & \hbhalf & \hbhalf \\
    \citet{deng2025selfimprovementmultimodallargelanguage}
    (\textit{MLLM Self-Improvement}) &
    arXiv 2025 & \hbhalf & \hbhalf & \hbnone & \hbfull \\
    \citet{yang2026selfimprovementlargelanguagemodels}
    (\textit{Self-Improvement}) &
    arXiv 2026 & \hbhalf & \hbhalf & \hbnone & \hbhalf \\
    \textbf{This survey} (\textit{Strict UPT}) &
    -- & \hbfull & \hbfull & \hbfull & \hbfull \\
    \bottomrule
  \end{tabularx}
  \captionsetup{hypcap=false}
  \captionof{table}{Scope dimensions of this survey and the closest existing
    surveys, grouped by publication year. \hbfull{}~explicit focus;
    \hbhalf{}~mixed coverage; \hbnone{}~outside the primary scope.
    \emph{Real update}: every surveyed
    method must induce an explicit parameter or state update.
    \emph{No external signal}: the update loop uses no ground-truth
    answers, verifier feedback, human labels, or stronger-teacher
    labels. \emph{Update-object taxonomy}: methods are organized by
    the internal object the gradient consumes. \emph{MLLM
    coverage}: vision-language or other multimodal foundation
  models are included.}
  \label{tab:survey-comparison}
\end{strip}

\section{Comparison with Existing Surveys}
\label{app:survey-comparison}

Table~\ref{tab:survey-comparison} places this survey against the
closest surveys along four scope dimensions. Prior work surveys
self-feedback and self-improvement
\citep{tao2024selfevolution,liang2024internalconsistencyselffeedbacklarge,
deng2025selfimprovementmultimodallargelanguage,
yang2026selfimprovementlargelanguagemodels}, broad LLM post-training
and reward learning
\citep{kumar2025posttraining,tie2025surveyposttraininglargelanguage,
wu2025sailingstarssurveyreward}, test-time adaptation
\citep{liang2025ttasurvey}, and
reinforced reasoning
\citep{xu2025reasoningmodels,chen2025longcotsurvey}. These scopes
overlap with parts of strict UPT, but none makes all four dimensions
a joint inclusion rule. Our distinctive unit of analysis is the
internal object consumed by an explicit update under the
no-external-signal boundary, across text and multimodal models.

\section{Use of AI Assistants}
\label{app:ai-assistants}
AI assistants were used for language polishing, including improving
grammar, clarity, and phrasing. All revisions were reviewed and verified
by the authors, who take full responsibility for the final manuscript.

\end{document}